\documentclass[journal]{IEEEtran}
\usepackage{amsmath,amsfonts}
\usepackage{algorithmic}
\usepackage{algorithm}
\usepackage{array}
\usepackage{textcomp}
\usepackage{stfloats}
\usepackage{url}
\usepackage{verbatim}
\usepackage{graphicx}
\usepackage{cite}
\usepackage{color}
\usepackage{booktabs}
\usepackage{threeparttable}
\usepackage{multirow}
\usepackage{makecell}
\usepackage{float}
\usepackage{siunitx}

\usepackage{orcidlink}
\hypersetup{hidelinks}
\newcommand{\etal}{\textit{et al.}}
\newcommand{\hhline}{\noalign{\vskip 1pt}\hline\noalign{\vskip 1pt}}

\makeatletter
\let\MYcaption\@makecaption
\makeatother
\usepackage[font=scriptsize]{subcaption}
\makeatletter
\let\@makecaption\MYcaption
\makeatother

\begin{document}

\title{Phase-aggregated Dual-branch Network for Efficient Fingerprint Dense Registration}

\author{Xiongjun Guan$^{\orcidlink{0000-0001-8887-3735}}$, 
	Jianjiang Feng$^{\orcidlink{0000-0003-4971-6707}}$, ~\IEEEmembership{Member, IEEE}, 
	and Jie Zhou$^{\orcidlink{0000-0001-7701-234X}}$, ~\IEEEmembership{Senior Member, IEEE}
	
	\IEEEcompsocitemizethanks{\IEEEcompsocthanksitem
    This work was supported in part by the National Natural Science Foundation of China under Grant 61976121 and 62376132.
	The authors are with Department of Automation, Tsinghua University, Beijing 100084, China (e-mail: gxj21@mails.tsinghua.edu.cn; jfeng@tsinghua.edu.cn; jzhou@tsinghua.edu.cn).
	}
}

\markboth{Journal of \LaTeX\ Class Files,~Vol.~14, No.~8, August~2021}%
{Guan \MakeLowercase{\textit{et al.}}: Phase-aggregated Dual-branch Network for Efficient Fingerprint Dense Registration}


\maketitle

\begin{abstract}
Fingerprint dense registration aims to finely align fingerprint pairs at the pixel level, thereby reducing intra-class differences caused by distortion.
Unfortunately, traditional methods exhibited subpar performance when dealing with low-quality fingerprints while suffering from slow inference speed.
Although deep learning based approaches shows significant improvement in these aspects, their registration accuracy is still unsatisfactory.
In this paper, we propose a Phase-aggregated Dual-branch Registration Network (PDRNet) to aggregate the advantages of both types of methods.
%{\color{blue}A dual-branch structure with multi-stage interactions is introduced between correlation information at high resolution and texture feature at low resolution, to perceive local fine differences while ensuring global stability.}
A dual-branch structure with multi-stage interactions is introduced between correlation information at high resolution and texture feature at low resolution, to perceive local fine differences while ensuring global stability.
Extensive experiments are conducted on more comprehensive databases compared to previous works.
Experimental results demonstrate that our method reaches the state-of-the-art registration performance in terms of accuracy and robustness, while maintaining considerable competitiveness in efficiency.

\end{abstract}


\section{Introduction}
\IEEEPARstart{B}{iometric} systems identify individuals through anatomical or behavioral characteristics.
Over the past few decades, many biometric traits have been suggested such as fingerprint, face, iris, vein, signature, and voice.
Among them, fingerprint is widely used in civil and criminal applications due to its high distinguishability and stability.
Although researchers have proposed many fingerprint matching algorithms, it is still a challenging task to deal with intra-class differences resulting from fingerprint distortion \cite{maltoni2022handbook}.

Fingerprint dense registration algorithms aims to establish dense correspondences at the pixel-level between fingerprint pairs.
Ridge curves of the registered fingerprint pairs can be strictly aligned, thus significantly reducing the negative impact of skin distortion and improving matching accuracy \cite{si2017dense}. 
Additionally, multiple fingerprints of the same finger can be mosaicked into a complete image after registration, thereby reducing storage requirements and expanding the effective matching area (especially important for small fingerprints) \cite{cui2021dense}.
Moreover, the registration results can serve as ground truth for certain tasks, such as distortion rectification \cite{guan2023regression} or modality transformation \cite{grosz2022c2cl}, which can conveniently obtain precise real data instead of synthetic.

\begin{figure}[!t]
	\centering
	\includegraphics[width=.85\linewidth]{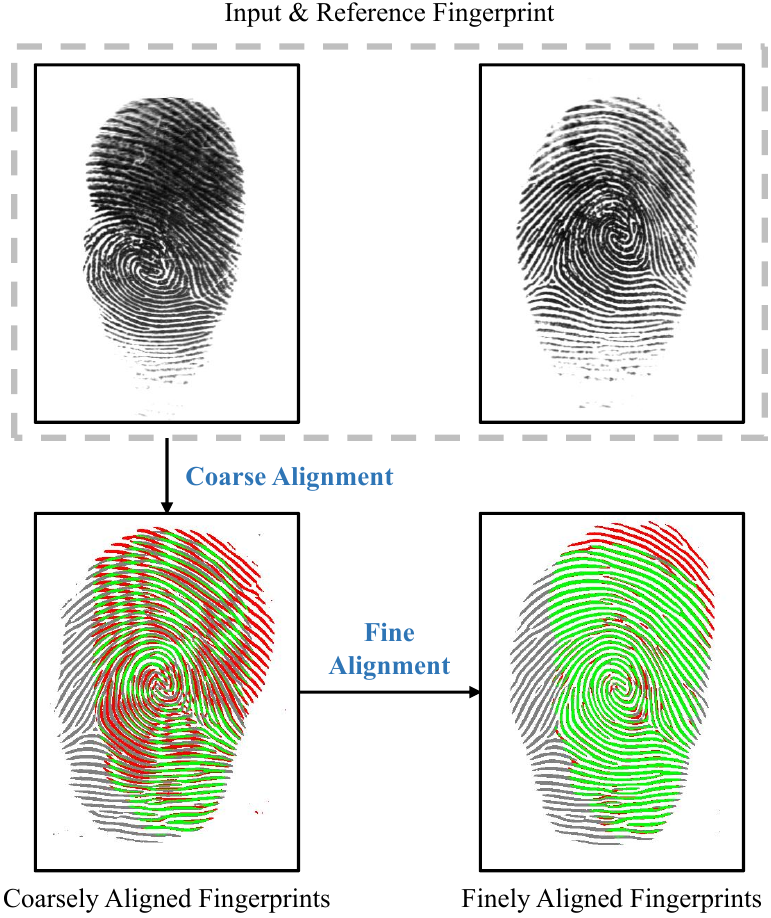}
	\caption{Flowchart of proposed two-step fingerprint dense registration. Green areas indicate overlapping ridges, gray and red indicate non-overlapping ridges of the two fingerprints respectively.}
	\label{fig:registration_task}
\end{figure}

Early conventional fingerprint registration algorithms commonly employ spatial transformation models based on matched minutiae pairs \cite{bazen2003fingerprint}, \cite{tico2003fingerprint}, \cite{ross2005deformable}. 
However, these methods cannot achieve precise alignment at ridge level and exhibit instability in regions with sparse minutiae or low quality.
In subsequent studies, it is developed and subdivided into two steps: first, a coarse alignment is conducted using rigid transformation or Thin Plate Spline transformation (TPS), then a fine alignment is performed to make local precise adjustments on ridge curves.
Although traditional dense registration methods \cite{si2017dense}, \cite{cui2018phase}, \cite{lan2020preregistration} are theoretically able to align local ridges strictly, they are very slow and susceptible to noise in practice.
On the other hand, dense registration methods based on convolutional neural networks \cite{cui2019dense}, \cite{cui2021dense} shows significant advantages in efficiency and robustness, but still needs to be improved in accuracy.
Therefore, we are motivated to design an innovative fingerprint dense registration method to integrate the advantages of both types of fingerprint dense registration solutions.
Phase features \cite{cui2018phase}, which can be easily obtained through 2D complex Gabor filters and accurately describe the offset of ridge points, meets our demands conceptually.
Inspired by multi branch networks \cite{sun2019deep}, \cite{yu2021bisenetv2},\cite{kumaar2021cabinet},\cite{pan2023deep}, we explicitly introduce the phase feature and make it as a branch of proposed network to better express the dense mapping relationship between fingerprint pairs.

In this paper, we propose a Phase-aggregated Dual-branch Registration Network (PDRNet) to combine the strengths of traditional fingerprint dense registration methods and deep learning.
The flowchart of our registration algorithm is shown in Fig. \ref{fig:registration_task}. 
For a fingerprint pair, named input fingerprint and reference fingerprint respectively, we first apply the TPS transformation based on matching minutiae for coarsely alignment (a robust rigid transformation method is used as the guarantee), then estimate the dense deformation field through PDRNet and perform fine alignment.
In contrast to previous deep learning methods that simply use pass-through \cite{cui2019dense} or encoder-decoder structures \cite{cui2021dense}, PDRNet introduces a dual-branch structure with multi-stage interactions between correlation features (specifically referred to phase in this paper) at high resolution and texture features at low resolution, to perceive local fine differences while ensuring global stability.
Moreover, the deformation field is estimated based on probability distribution of discretized intervals instead of previous direct regression methods, inspired by \cite{gidaris2016locnet},\cite{tang2017fingernet},\cite{doosti2020hopenet}.

We conducted extensive experiments on databases containing different types of fingerprint impressions, including different sensing technologies (optical, thermal wiped, latent, non-contact) and different skin conditions (normal, dry, wet, distorted, incomplete, aged). 
Experimental results demonstrate that the proposed algorithm achieves state-of-the-art registration performance, while also having great competitiveness in terms of model size and inference speed.

The main contributions of this work can be summarized as follows:
\begin{itemize}
	\item Phase features that perform well in traditional fingerprint registration \cite{cui2018phase} are introduced into the proposed convolutional neural network, enabling it to perceive more refined correlation information;
	\item We introduce a multi-stage feature interaction mechanism to handle dual-branch information, which can combine the robustness of low-resolution texture features and the sensitivity of high-resolution correlation features.
	\item The numerical regression task in previous works is transformed into interval classification, further improving the performance of deformation field estimation.
	\item Extensive experiments are conducted on more abundant datasets compared to previous studies.
	Comprehensive evaluation of registration and matching performance is performed across multiple fingerprints impressions.
\end{itemize}

The paper is organized as follows. 
Section \ref{sec:related_work} reviews the related works. 
Section \ref{sec:method} introduces the proposed dense registration algorithm.
Section \ref{sec:dataset} describes experiment datasets we used in training and evaluation, and implementation details.
Section \ref{sec:experiment} presents the experimental results and discussions.
Finally, we make conclusions in Section \ref{sec:conclusion}.

\section{Related Work}\label{sec:related_work}
Briefly, fingerprint registration aims to establish the correspondence between input fingerprints and reference fingerprints, and then transform the fingerprint pairs to align them as closely as possible.
According to the transformation model, fingerprint registration can be divided into two categories: rigid transformation and elastic registration.
From the distribution density of control points, elastic registration can be further subdivided into sparse or dense registration.

\subsection{Fingerprint Rigid Transformation} \label{subsec:rigid_transformation}
Early fingerprint registration studies mostly utilize rigid transformations, which only consider the relative translation, rotation, or scaling relationships between image pairs.
These algorithms commonly estimate transformation parameters by minimizing the projection error of matching minutiae \cite{ratha1996real},\cite{tico2003fingerprint},\cite{moon2004template},\cite{zhu2005fingerprint}, orientation field \cite{yager2004evaluation},\cite{liu2006fingerprint} or image correlation \cite{bazen2000correlation},\cite{hatano2002fingerprint}. 
%{\color{blue}On the other hand, some researchers recently utilize Spatial Transformer Network (STN) combined with subsequent recognition tasks to estimate affine transformation parameters of each input fingerprint \cite{engelsma2021learning,grosz2023afrnet}, or use more complex networks to directly estimate the fingerprint pose \cite{yin2021joint,duan2023estimating}. Although these algorithms are applied to a single fingerprint, they have a certain ability to roughly align fingerprint pairs.}
On the other hand, some researchers recently utilize Spatial Transformer Network (STN) combined with subsequent recognition tasks to estimate affine transformation parameters of each input fingerprint \cite{engelsma2021learning,grosz2023afrnet}, or use more complex networks to directly estimate the fingerprint pose \cite{yin2021joint,duan2023estimating}. Although these algorithms are applied to a single fingerprint, they have a certain ability to roughly align fingerprint pairs.
Obviously, these models lack the capability to handle the nonlinear distortion of fingerprints.

\subsection{Fingerprint Sparse Registration} \label{subsec:sparse_registration}
Sparse registration models calculates transformation parameters based on the position of control points.
Among them, TPS based models that use minutiae correspondences are widely used \cite{almansa2000fingerprint}, \cite{bazen2003fingerprint},\cite{ross2005deformable},\cite{ross2006image} because it can align the extracted stable feature points while ensuring smooth mapping relationships.
However, these methods cannot accurately register areas without or far from minutiae.
Researchers have introduced various extended features, such as the orientation field \cite{chen2009reconstructing}, period map \cite{wan2006fingerprint}, ridge curve \cite{feng2006fingerprint} and network descriptor \cite{tang2017fingernet},\cite{cao2020end}, to improve the accuracy of feature point extraction and matching.
Nevertheless, these advancements do not fully address the issue of unreliable results in low-quality areas.
Furthermore, it poses a significant challenge to establish dense correspondence between featureless ridge points solely based on sparsely distributed control points in space.

\subsection{Fingerprint Dense Registration}
Fingerprint dense registration aims to establish pixel by pixel correspondence between fingerprint pairs.
Researchers typically use methods introduced in Section \ref{subsec:rigid_transformation} or Section \ref{subsec:sparse_registration} to roughly align fingerprints and then make fine adjustments in the following stage.

Si \etal \cite{si2017dense} proposed a dense registration method based on block-based image correlation and Markov optimization, which performs better than sparse registration methods.
Cui \etal \cite{cui2018phase} introduced the concept of phase demodulation into fingerprint registration, resulting in substantial improvements in both speed and accuracy.
Lan \etal \cite{lan2020preregistration} also proposed a simple but effective registration algorithm, which implements iterative optimization based on correlation and orientation field, but sometimes be unstable on images with significant grayscale variations.
This kind of methods all depend on manually designed image features and are susceptible to areas with low quality or large distortion. 
In addition, their optimized ways of iteration or traversal are computationally expensive.

Convolutional neural networks are used to regress pixel-wise displacements in methods of Cui \etal \cite{cui2019dense},\cite{cui2021dense}.
Compared with traditional algorithms, deep learning based methods show desirable advantages in terms of robustness and efficiency.
However, their works simply connected a siamese network with straight network \cite{cui2019dense} or encoder-decoder \cite{cui2021dense} without incorporating additional constraints on the learned features, which might not be sensitive enough to differences in fingerprint pairs.

\begin{figure*}[!t]
	\begin{center}
		\includegraphics[width=1\linewidth]{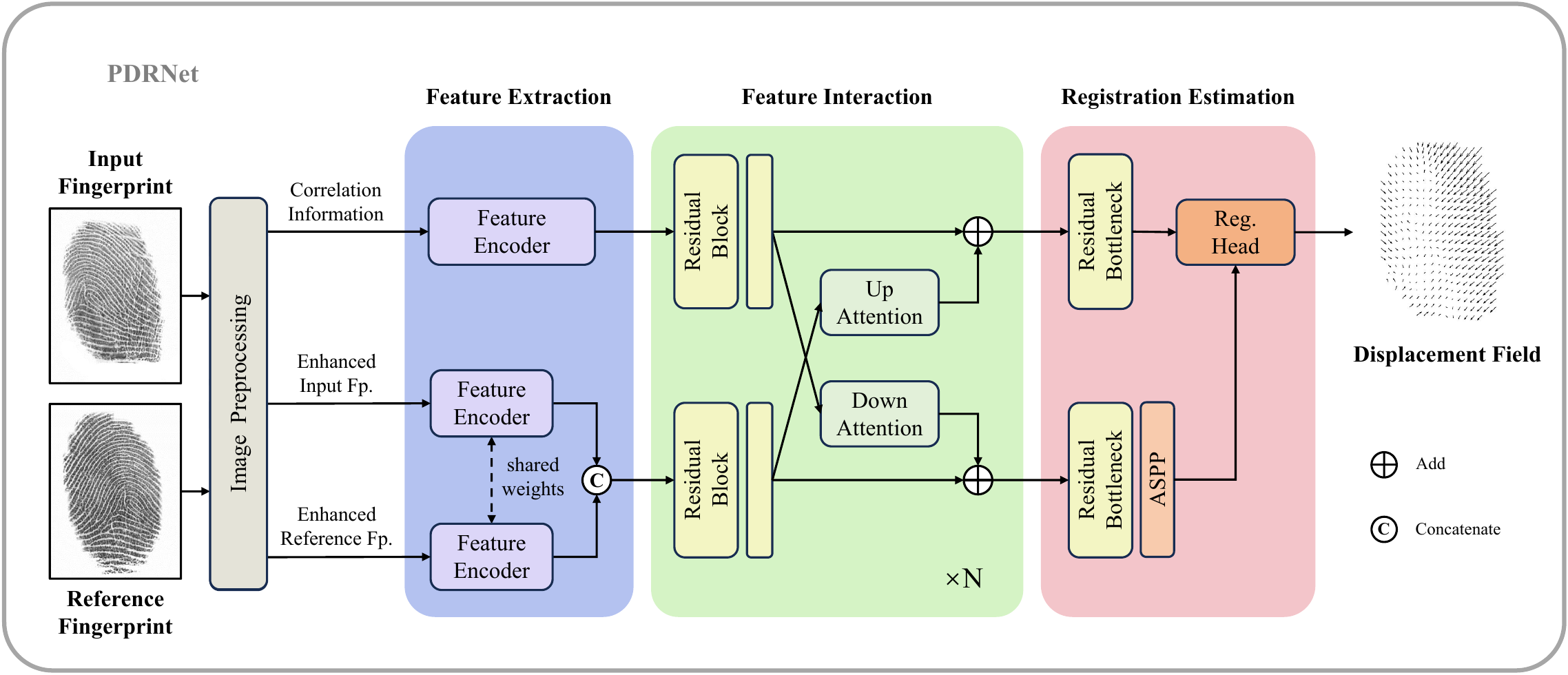}
	\end{center}
	\caption{An overview of our dense deformation estimation network. `Fp.' and `Reg.' are the abbreviations of `fingerprint' and `registration' respectively. The network includes two encoder branches for extracting features of texture and correlation information, a multi-stage interaction module for fusing multi-scale and multi-semantic features, and an registration estimation module to predict the pixel-wise deformation field. The specific details of block architectures and image preprocessing flow are shown in Fig. \ref{fig:block_architecture} and Fig. \ref{fig:preprocess_flowchart} respectively.  }
	\label{fig:network}
\end{figure*}

\begin{figure}[!t]
	\centering
	\includegraphics[width=1\linewidth]{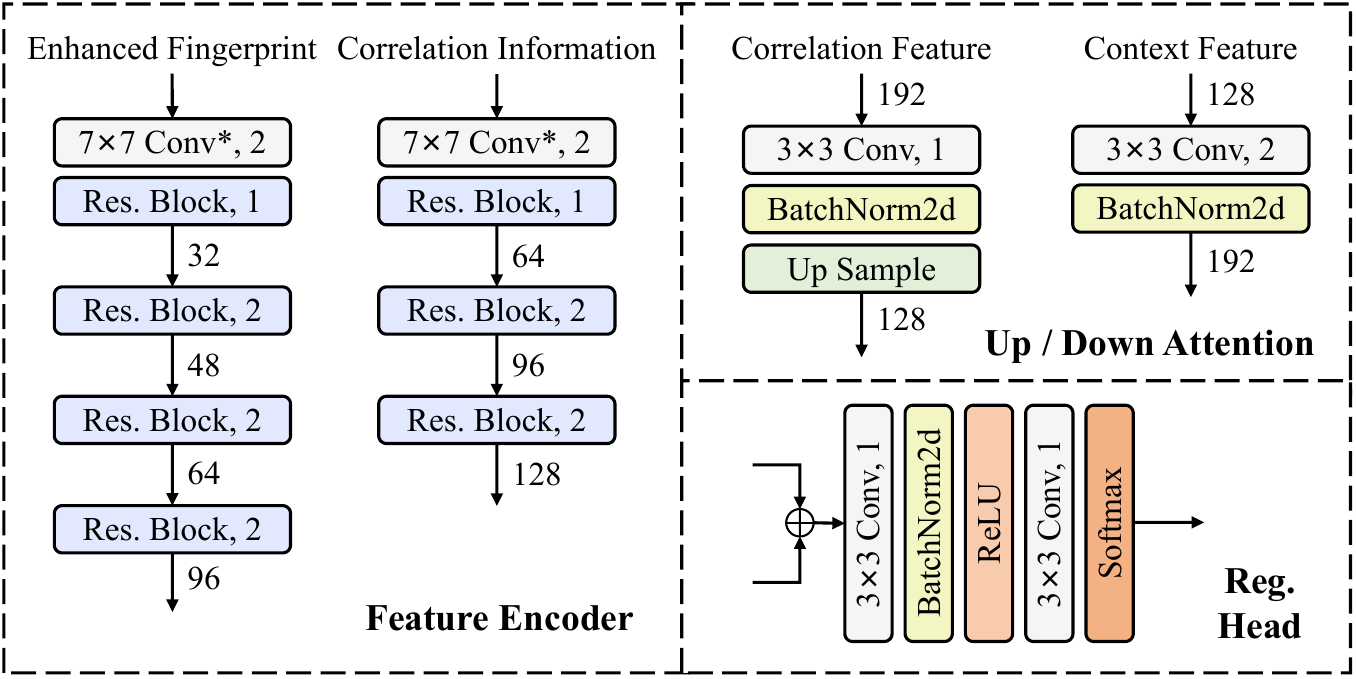}
	\caption{The specific architecture of blocks utilized in the proposed fingerprint dense registration network. `*' indicates that this layer connects convolution, batch normalization and ReLU in series. Numbers on the right side of each module and arrows represent the stride and current channel number respectively.}
	\label{fig:block_architecture}
\end{figure}

\section{Method} \label{sec:method}
In this paper, we estimate the pixel-level displacement fields between fingerprint pairs in two stages to perform dense registration.
Fig. \ref{fig:registration_task} gives the complete flowchart of the proposed algorithm.
For a pair of fingerprints, our algorithm first apply the TPS model based on matching minutiae for coarse alignment.
A robust rigid transformation method based on orientation and period map is used as a fallback. 
The preliminarily aligned fingerprints are then preprocessed as enhancement images and correlation information (determined as phase in final approach) and input into the proposed network, whose structure is shown in Fig. \ref{fig:network}, so as to obtain a dense displacement field from the input fingerprint to the reference fingerprint. 
Finally, the input image is moved pixel by pixel according to the network prediction result to approximate the reference fingerprint.

\subsection{Coarse Alignment} \label{subsec:coarse_alignment}
Same as previous works \cite{si2017dense},\cite{cui2018phase},\cite{cui2019dense},\cite{cui2021dense}, we perform coarse registration based on fingerprint minutiae.
VeriFinger \cite{VeriFinger}, a widely used commercial software, is utilized for minutiae extraction and matching. The paired points are subsequently used as control points for TPS transformation to coarsely align the input and reference fingerprints.
It should be noted that the TPS model may not be reliable when the number of paired minutiae is small.
Therefore, we switch to conduct a global search for the optimal rigid transformation based on orientation and period maps in order to obtain a more accurate alignment result, due to their advantages in registration robustness in these cases \cite{krish2015pre}.
The function for parameter search is defined as
\begin{equation}
	\begin{aligned}
		\underset{x, y, \theta}{\operatorname{argmax}} \; \|  & \text{OriDiff} \left({O}_{\mathrm{R}}, {O}_{\mathrm{I}}(x, y, \theta)\right) \leq \theta_{\mathrm{t}} \; \& \\
		& \text{PedDiff}\left({P}_{\mathrm{R}}, {P}_{\mathrm{I}}(x, y, \theta)\right) \leq p_{\mathrm{t}} \; \|_0 \;,
	\end{aligned}
\end{equation}
where $x$, $y$ and $\theta$ represent the translation and rotation parameters, $O$ is the orientation map, $P$ is the period map, subscripts $\mathrm{I}$ and $\mathrm{R}$ represent the input and reference fingerprint respectively, functions $\text{OriDiff}\left(\right)$ and $\text{PedDiff}\left(\right)$ calculate the difference of orientation and period maps at the corresponding location.
Fixed thresholds $\theta_{\mathrm{t}}$ and $p_{\mathrm{t}}$ are set to $10^{\circ}$ and $1$ pixel.
Both orientation and period map are sampled on blocks of $8 \times 8$ blocks.
We perform the above rigid transformation when paired minutiae are less than $4$, otherwise use TPS-based sparse elastic registration.

\subsection{Image Preprocessing} \label{subsec:image_preprocessing}

\begin{figure*}[!t]
	\centering
	\includegraphics[width=1\linewidth]{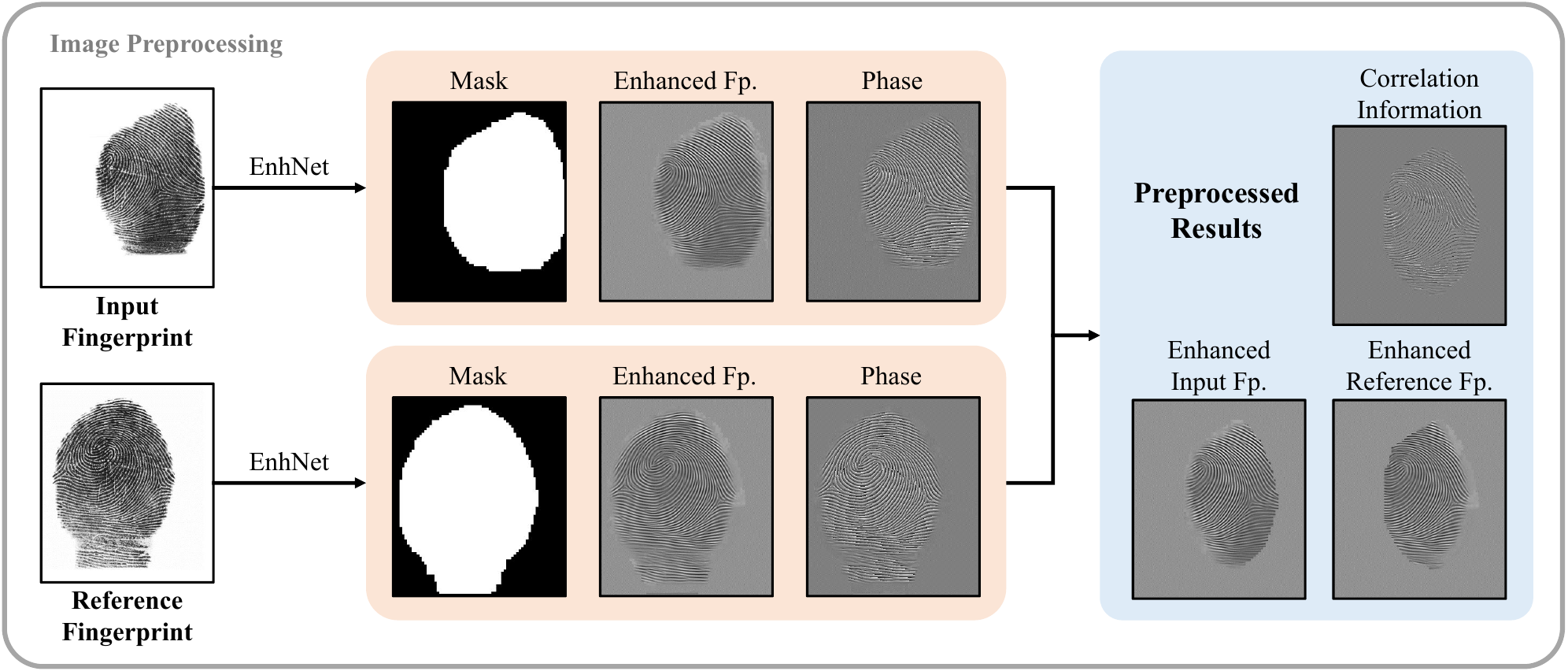}
	\caption{A visual example of image preprocessing. Red rectangles represent the intermediate results of fingerprint enhancement. Blue rectangle represents the final result of image preprocessing, which corresponds to the output of the same module in Fig. \ref{fig:network}. The structure of `EnhNet' is shown in Fig. \ref{fig:network_preprocess}.}
	\label{fig:preprocess_flowchart}
\end{figure*}

\begin{figure}[!t]
	\centering
	\includegraphics[width=0.9\linewidth]{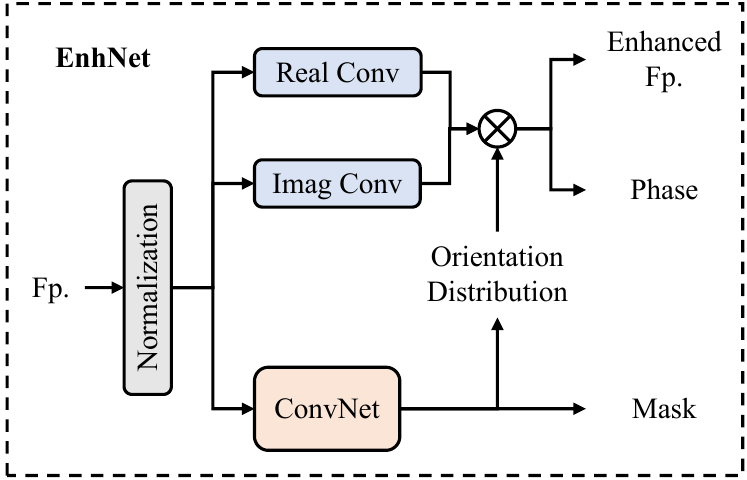}
	\caption{A simplified schematic of fingerprint enhancement network. The main structure is separated from FingerNet \cite{tang2017fingernet}, and only the output is adjusted according to Equation \ref{eq:enh_phase}.}
	\label{fig:network_preprocess}
\end{figure}

Image quality of fingerprints greatly affects the performance of subsequent algorithms. 
Researchers have proposed several enhancement methods in the preprocessing stage to improve the clarity of ridge structures.
Among them, context filtering in the spatial domain based on Gabor filters \cite{daugman1985uncertainly} can effectively remove undesired noise while retaining the true ridge structure, which is currently one of the most popular technologies \cite{maltoni2022handbook}.
On the other hand, phase features can also be calculated through Gabor filter banks \cite{tang2017fingernet}. 
In other words, we can naturally obtain phase feature from the intermediate results of enhancement without additional complex operations.
Therefore, we integrate fingerprint enhancement and phase feature extraction in the same module.
For convenience, this module is shown in Fig. \ref{fig:network} as the first module of the proposed network, but it can also be executed independently of PDRNet.

The 2-D Gabor filter combines Gaussian waves and sine waves.
Its expression in the complex domain after adjustment for fingerprints is:
\begin{equation}
	G(x, y)=  \exp \left(-\left(\frac{x^{\prime 2}}{2 \sigma_x^2}+\frac{y^{\prime 2}}{2 \sigma_y^2}\right)\right)   \exp \left(i \cdot 2 \pi f_0 x^{\prime}\right) ,
\end{equation}
where $\sigma_x$ and $\sigma_y$ are the  standard deviation of corresponding direction,  $x^{\prime}$ and $y^{\prime}$ are the original coordinates $x$ and $y$ rotated by angle $\theta$, $f_0$ is the filtering frequency.
The $\theta$ and $f_0$ parameters should be adjusted according to the orientation and period of local ridges in practical.
Let $Z(x,y)$ denote the filtered result with $G(x,y)$, the enhanced fingerprint $E$ and phase $\phi$ of the original image can be calculated as
\begin{equation}
	\begin{aligned}
		E & = \text{Norm}\left( \operatorname{Re} \left[ Z \right]\right), \\
		\phi(x, y) & =\operatorname{atan2}\left(\operatorname{Re}\left[Z(x, y)\right], \operatorname{Im}\left[Z(x, y)\right]\right),
	\end{aligned}
\label{eq:enh_phase}
\end{equation}
where $\text{Norm}\left( \right)$ is to perform global normalization, $\operatorname{Re}\left[Z\right]$ and $\operatorname{Re}\left[Z\right]$ are the real and imaginary components of the complex signal $Z$ respectively.

A partial structure of FingerNet \cite{tang2017fingernet} is isolated and adjusted (according to Equation \ref{eq:enh_phase}) for this task because it can obtain the above features quickly and accurately.
As show in Fig. \ref{fig:network_preprocess}, this network converts the Gabor filter bank as a set of convolution kernels with fixed parameters and estimates the orientation field through another branch to select appropriate filtering results. The mask of the fingerprint is also estimated.
Other enhancement methods based on Gabor filtering can also be used to obtain similar results.

In addition to enhancement images $E$ and phases $\phi$, mask $M$ is also utilized to reduce interference in non-overlapping areas. For the input fingerprint $I$ and reference fingerprint $R$, the final results of image preprocessing is
\begin{equation}
	\begin{aligned}
		E^{\prime}_{\mathrm{I}} &= E_{\mathrm{I}} \cdot M_{\mathrm{I}}, \quad E^{\prime}_{\mathrm{R}} = E_{\mathrm{R}} \cdot M_{\mathrm{R}}, \\
		\psi  &= \left(\phi_{\mathrm{I}}-\phi_{\mathrm{R}}\right) \cdot M_{\mathrm{I}}\cdot M_{\mathrm{R}},
	\end{aligned}
	\label{eq:enhancement}
\end{equation}
where $\psi$ represents the correlation information, $E^{\prime}_{\mathrm{I}}$ and $E^{\prime}_{\mathrm{R}}$ represent the enhanced input and reference fingerprint respectively.
An example is given in Fig. \ref{fig:preprocess_flowchart}, which visualizes the intermediate and final results in the image preprocessing stage.

\subsection{Network for Dense Displacement Field Estimation}
For a pair of fingerprints, our proposed network predicts the dense deformation field from the input fingerprint to the reference fingerprint.
A two-branch structure is introduced, which extracts features of different semantics at different resolutions respectively and performs information interaction in multiple stages, to understand local ridge differences finely and robustly.
In this paper, phase feature are utilized as correlation information because it is easy to obtain and can accurately describe the displacement relationship at pixel level \cite{cui2018phase}.
The complete structure of PDRNet is given in Fig. \ref{fig:network}, which can be divided into four parts: image preprocessing, feature extraction, feature interaction and registration estimation. 
In Section \ref{subsec:image_preprocessing}, we have described how to perform image preprocessing on fingerprint pairs. 
The following mainly introduces the design motivation and specific implementation of remaining architectures and loss functions of PDRNet.

\subsubsection{Feature Extraction}

\begin{figure}[!t]
	\centering
	\includegraphics[width=1\linewidth]{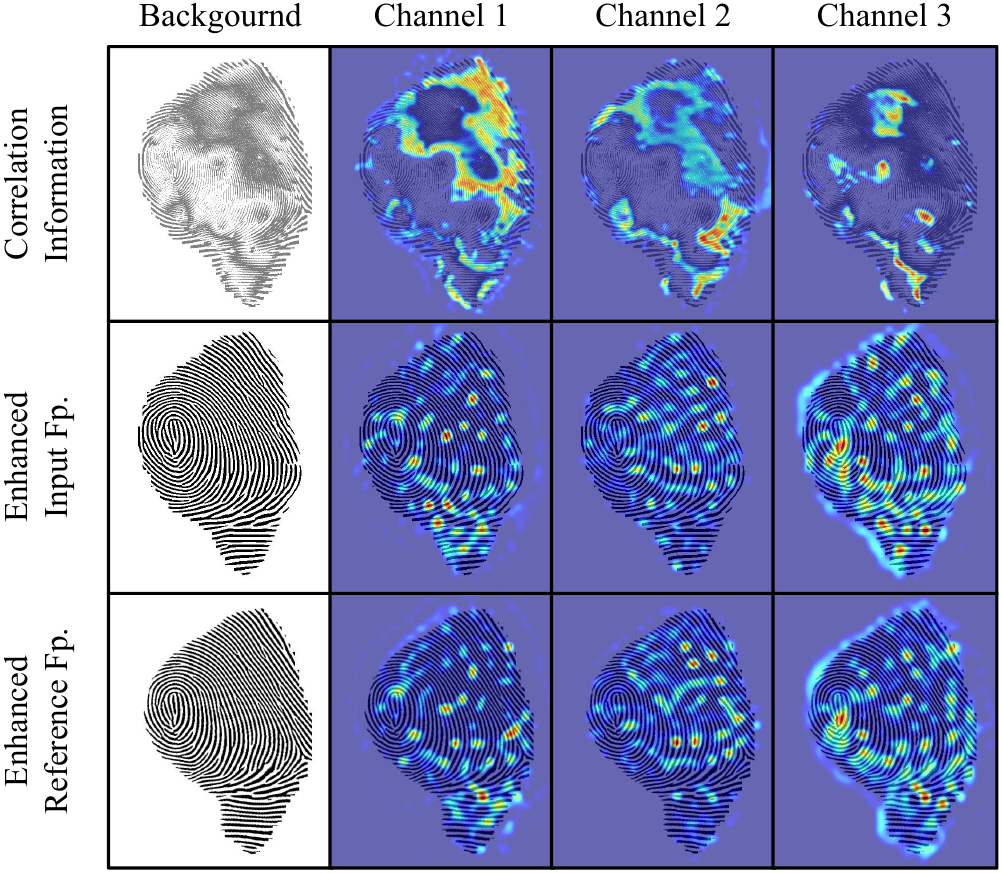}
	\caption{Examples of heatmap visualized from extracted features, which reflects the interesting areas of encoders from different branches. Each row, from left to right, is the background of subsequent images and attention maps from three channels output by the respective encoders. The background from bottom to top is the binary image of two fingerprints and the XOR result between them.}
	\label{fig:heatmap}
\end{figure}

A natural idea is to extract texture features directly from images.
In work \cite{cui2019dense} and \cite{cui2021dense}, a siamese network with shared parameters was applied, which first extracts the respective texture features from each image separately and then merges and analyzes them.
These implementations are similar to the control points based elastic registration in traditional methods. As analyzed in Section \ref{subsec:sparse_registration}, even if stable feature points can be extracted and matched, their number is not enough to establish a pixel-wise dense connections.
Therefore, we introduce correlation information that can accurately represent the displacement relationship and explicitly use it as network input, while retaining the previous disposal mode of texture features.
As shown in Fig. \ref{fig:block_architecture}, downsampled convolutions are stacked as feature encoders in order to balance registration accuracy and model parameters.
For features of correlation information, we only downsample them to $1/8$ size to retain richer spatial details. 
On the other hand, texture features are calculated using a deeper encoder ($1/16$ of the original size) to extract more abstract semantic information to increase the robustness of registration.

Heatmap of features extracted by the respective encoders are shown in Fig. \ref{fig:heatmap}.
It can be seen that the texture branch without additional constraints pays more attention to point-level features, and the high corresponding areas are sparsely distributed and relatively rough; while features inferred from correlation information pay more attention to regional properties, and the high-response areas are densely distributed and relatively fine.
This case strongly demonstrates that the proposed two branches effectively capture information with distinct semantics.
%{\color{blue}One reasonable explanation is that the input to texture feature branch is the respective enhanced result and there is no information exchange between enhanced fingerprint pairs during feature extraction, which will make the corresponding encoder tend to focus on stable features from a single image in the form of points (such as minutiae and singular points) and ignore low-texture areas that are highly repetitive or easily confused. On the other hand, the input to correlation information branch is the phase difference between fingerprint pairs, whose value of each pixel clearly characterizes the offset of ridge alignment \cite{cui2018phase}, resulting in a regional response in the feature distribution due to the spatial continuity of fingerprint deformation.}
One reasonable explanation is that the input to texture feature branch is the respective enhanced result and there is no information exchange between enhanced fingerprint pairs during feature extraction, which will make the corresponding encoder tend to focus on stable features from a single image in the form of points (such as minutiae and singular points) and ignore low-texture areas that are highly repetitive or easily confused. 
On the other hand, the input to correlation information branch is the phase difference between fingerprint pairs, whose value of each pixel clearly characterizes the offset of ridge alignment \cite{cui2018phase}, resulting in a regional response in the feature distribution due to the spatial continuity of fingerprint deformation.
\subsubsection{Feature Interaction}
Previous works \cite{cui2019dense},\cite{cui2021dense} have verified that networks can establish the relationship between fingerprint pairs from extracted texture features.
Furthermore, the studies conducted by \cite{spoorthi2020phasenet2} and \cite{cui2018phase} proved that networks have the ability to unwrap phase and thus resolve the displacement between fingerpints.
Therefore, it is theoretically feasible to analyze the displacement through dual-branch features in PDRNet.
%{\color{blue}Inspired by other multi-branch network structure designs \cite{sun2019deep},\cite{yu2021bisenetv2},\cite{kumaar2021cabinet},\cite{pan2023deep} and multi-stage feature iteration strategies \cite{hur2019iterative},\cite{teed2020raft}, we introduce a bilateral structure to analyze and interact the information from two branches in several iterations.}
Inspired by other multi-branch network structure designs \cite{sun2019deep},\cite{yu2021bisenetv2},\cite{kumaar2021cabinet},\cite{pan2023deep} and multi-stage feature iteration strategies \cite{hur2019iterative},\cite{teed2020raft}, we introduce a bilateral structure to analyze and interact the information from two branches in several iterations.
Residual blocks \cite{he2016deep} are exploited to further express features and avoid gradient problems, while simple convolution and sampling layers are used for information exchange between branches.
It should be noted that the shape of each feature is not changed in order to facilitate arbitrary adjustment of iterations.

\subsubsection{Registration Estimation} \label{subsubsec:registration_estimation}
The iterated results are first fed into a residual bottleneck \cite{he2016deep} to further integrate features, where the channels are doubled to carry denser information.
The ASPP module \cite{chen2018encoder} (with dilated ratio of 1,2,4 and average pooling) is connected next to capture texture information at multiple scales and fuse features. 
Here we perform such a transformation flow because phase is used as correlation information in this paper, which can be mapped to displacement more directly and interpretably than texture features \cite{cui2018phase}.
Fig. \ref{fig:block_architecture} shows the specific structure of the registration head.
Inspired by \cite{gidaris2016locnet},\cite{tang2017fingernet},\cite{doosti2020hopenet}, we use interval classification to express the deformation field, which can limit the range to avoid unreasonable output and characterize the numerical distribution thus reduce the learning difficulty, instead of previous direct regression \cite{cui2019dense},\cite{cui2021dense}.
Let $p^t$ and $z^t$ represent the numerical value and predicted probability corresponding to $t$-th category, the final deformation field $D$ is calculated as
\begin{equation}
	D = \frac{1}{\sum_T p^t}\sum_T p^t z^t ,
	\label{eq:deformation}
\end{equation}
and subsequently interpolated to the full size.
Considering that large relative distortions are basically removed through the coarse alignment in Section \ref{subsec:coarse_alignment}, we set the value range of displacement in both directions to $\left[-30,30\right]$ pixels and divide it into $25$ equal intervals, that is, 50 channels are output with $1/8$ size of original image.
The deformation field in non-overlapping regions can be approximated by TPS interpolation based on estimated results in overlapping regions.

\subsubsection{Loss Function}
Considering that directly using 0-1 labels cannot accurately express values especially at the boundaries of intervals, we use Gaussian smoothing label strategy to generate the ground truth of class probability $p_\mathrm{gt}$ as 
\begin{equation}
	p_\mathrm{gt}^t \left(z\right) = \frac{1}{\sum_{t}{p_\mathrm{gt}^t\left(z\right)}} \exp \left(-\frac{z-z^t}{2 \sigma^2}\right) ,
	\label{eq:gt}
\end{equation}
where $z$ represents the real displacement, $z^t$ represents the discrete value corresponding to the $t$-th category. The variance $\sigma$ is set to $2.0$ empirically.
Focal loss \cite{lin2017focal} is applied to optimize these imbalanced multi-class distributions, which is defined as
\begin{equation}
	\begin{aligned}
		\mathcal{L}_{\mathrm{cla}} & =-\frac{1}{|M|} \sum_M \sum_{t=1}^T \alpha\left(1-q^t\right)^\gamma \log \left(q^t\right), \\
		q^t & =p_\mathrm{gt}^t p^t+\left(1-p_\mathrm{gt}^t\right)\left(1-p^t\right) ,
	\end{aligned}
\end{equation}
where $M$ is the mask of the common area of fingerprint pairs, $p^t$ is the probability that the deformation is inferred in $t$-th  category, $T$ is the total number of intervals.
Hyperparameters $\alpha$ and $\gamma$ are fixed to $1.0$ and $2.0$ respectively.
Another objective function $\mathcal{L}_{\mathrm{smo}}$ is also performed to constrain the spatial smoothness of probability distribution:
\begin{equation}
	\mathcal{L}_{\mathrm{smo}} = -\frac{1}{|M|} \sum_M \sum_{t=1}^T \left|\Delta p^t \right| , 
\end{equation}
where $\Delta$ is the standard Laplacian filter.
Overall, the complete loss function is
\begin{equation}
	\mathcal{L} = \mathcal{L}_{\mathrm{cla}} + \lambda \cdot \mathcal{L}_{\mathrm{smo}}.
\end{equation}
The weight $\lambda$ is fixed to $1.0$ in this paper.

\begin{table*}[!ht] 
	\caption{All fingerprint datasets used in experiments.}
	\label{tab:datasets}
	\vspace{-0.7cm}
	\begin{center}
		\begin{threeparttable}
			\begin{tabular}{p{.12\linewidth}<{\centering}*{1}{p{.24\linewidth}<{\centering}}*{1}{p{.14\linewidth}<{\centering}}*{1}{p{.24\linewidth}<{\centering}}*{1}{p{.14\linewidth}<{\centering}}}
				\toprule
				\multirow{2}{*}{Database} 	
				& \multicolumn{2}{c}{Hisign Latent}
				& TDF
				& THU Old \\
				\cmidrule(lr){2-3} \cmidrule(lr){4-4} \cmidrule(lr){5-5}
				& Rolled / Plain
				& Latent
				& Plain
				& Plain \\
				\midrule
				Image                     & \raisebox{-.5\height}{\includegraphics[height=.8in]{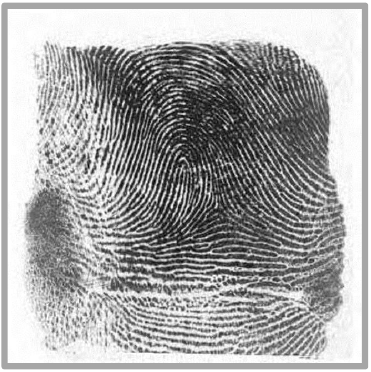}}~\raisebox{-.5\height}{\includegraphics[height=.8in]{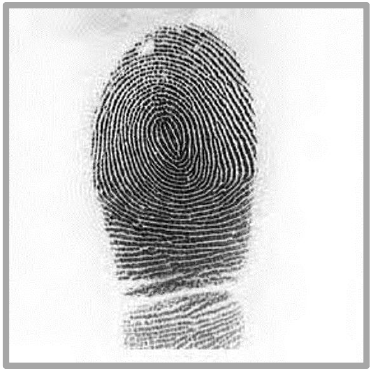}}
				& \raisebox{-.5\height}{\includegraphics[height=.8in]{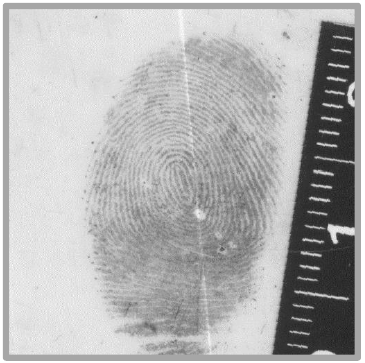}}
				& \raisebox{-.5\height}{\includegraphics[height=.8in]{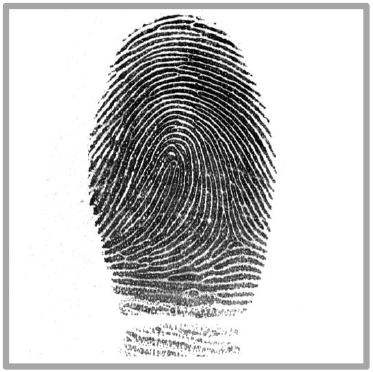}}~\raisebox{-.5\height}{\includegraphics[height=.8in]{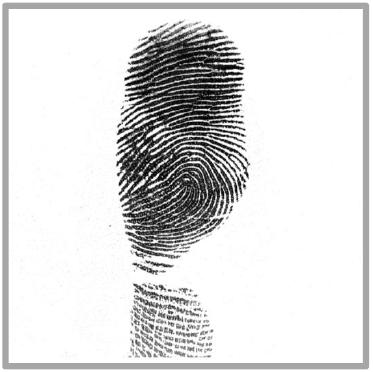}}
				& \raisebox{-.5\height}{\includegraphics[height=.8in]{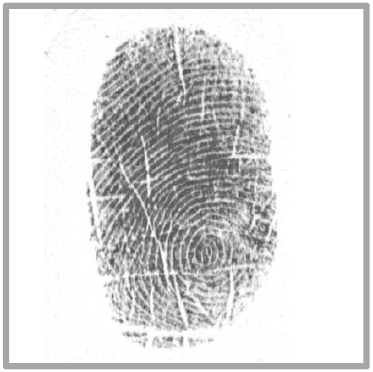}} \\
				\hhline
				Sensor
				& Inking / Optical                    
				& -
				& Optical
				& Optical \\
				\hhline
				Description              
				& \multicolumn{2}{c}{\makecell{10459 pairs \\ latent fingerprints from real crime scenes}}
				& \multicolumn{1}{c}{\makecell{320 videos \\ large distortion}}
				& \multicolumn{1}{c}{\makecell{136 fingers $\times$ 2\\ wrinkled and low quality}}\\
				\hhline
				Usage                     
				& \multicolumn{2}{c}{Training}
				& Training
				& \multicolumn{1}{c}{\makecell{Registration accuracy\\Matching performance}} \\
				\hhline
				Genuine Match
				& \multicolumn{2}{c}{$\backslash$}
				& $\backslash$
				& 136\tnote{a}\\
				\hhline
				Imposter Match
				& \multicolumn{2}{c}{$\backslash$}
				& $\backslash$ 
				& 9,180\tnote{b}\\
				\bottomrule
			\end{tabular}
			\vspace*{0.25mm}
			
			\begin{tabular}{p{.12\linewidth}<{\centering}*{1}{p{.36\linewidth}<{\centering}}*{1}{p{.135\linewidth}<{\centering}}*{2}{p{.135\linewidth}<{\centering}}}
				\toprule
				\multirow{2}{*}{Database} 
				& FVC2004 DB1\_A
				& FVC2004 DB3\_A
				& \multicolumn{2}{c}{NIST SD27} \\
				\cmidrule(lr){2-2} \cmidrule(lr){3-3} \cmidrule(lr){4-5}
				& Plain
				& Plain
				& Rolled
				& Latent \\
				\midrule
				Image                    
				& \raisebox{-.5\height}{\includegraphics[height=.8in]{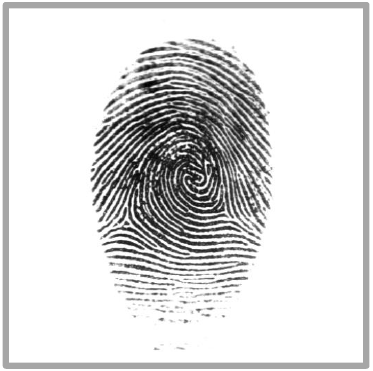}}~\raisebox{-.5\height}{\includegraphics[height=.8in]{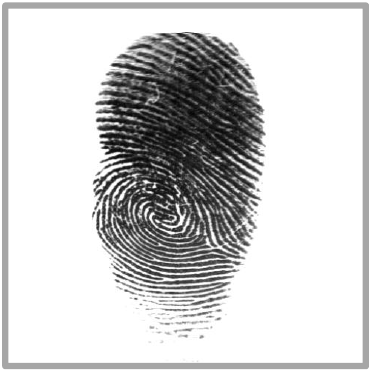}}~\raisebox{-.5\height}{\includegraphics[height=.8in]{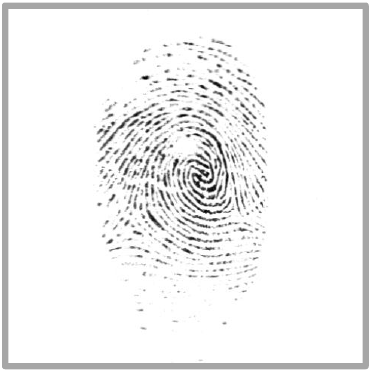}}
				& \raisebox{-.5\height}{\includegraphics[height=.8in]{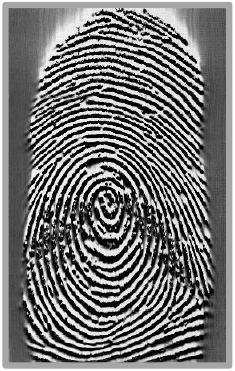}}
				& \raisebox{-.5\height}{\includegraphics[height=.8in]{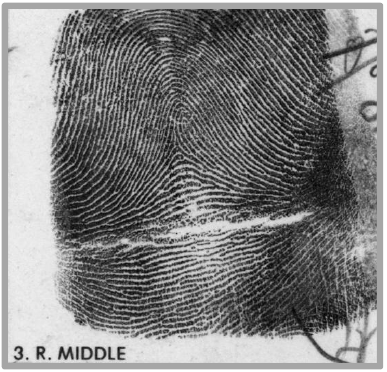}}
				& \raisebox{-.5\height}{\includegraphics[height=.8in]{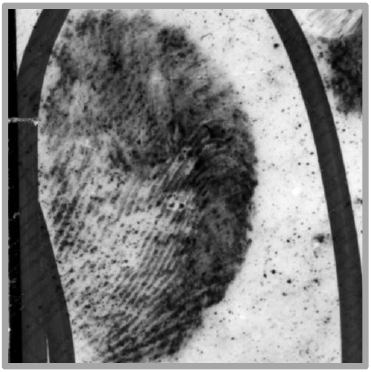}} \\
				\hhline
				Sensor                    
				& Optical
				& Thermal sweeping
				& Inking
				& - \\
				\hhline
				Description               
				& \multicolumn{1}{c}{\makecell{100 fingers $\times$ 8 \\ large distortion and various quality}}
				& \multicolumn{1}{c}{\makecell{100 fingers $\times$ 8 \\ large distortion}}
				& \multicolumn{2}{c}{\makecell{258 pairs \\ latent fingerprints from real crime scenes }} \\
				\hhline
				Usage                     
				& \multicolumn{1}{c}{\makecell{Registration accuracy\\Matching performance}}
				& \multicolumn{1}{c}{\makecell{Registration accuracy\\Matching performance}}
				& \multicolumn{2}{c}{\makecell{Matching performance}} \\
				\hhline
				Genuine Match
				& 2,800\tnote{a}
				& 2,800\tnote{a}
				& \multicolumn{2}{c}{258\tnote{a}}\\
				\hhline
				Imposter Match
				& 4,950\tnote{b}
				& 4,950\tnote{b}
				& \multicolumn{2}{c}{66,306\tnote{c}}\\
				\bottomrule
			\end{tabular}
			\vspace*{0.25mm}
			
			\begin{tabular}{p{.12\linewidth}<{\centering}*{1}{p{.335\linewidth}<{\centering}}*{1}{p{.12\linewidth}<{\centering}}*{1}{p{.335\linewidth}<{\centering}}}
				\toprule
				\multirow{2}{*}{Database} 
				& Hisign MPF
				& \multicolumn{2}{c}{Hisign C2CL} \\
				\cmidrule(lr){2-2} \cmidrule(lr){3-4}
				& Plain
				& Contact-based (C)
				& Contactless (CL) \\
				\midrule
				Image                    
				&
				\raisebox{-.5\height}{\includegraphics[height=.8in]{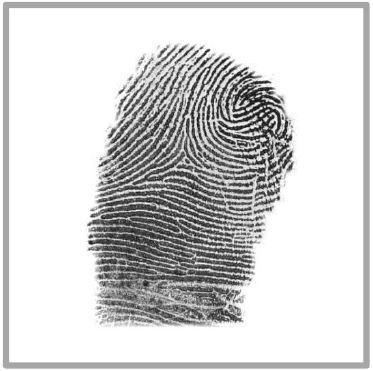}}~ \raisebox{-.5\height}{\includegraphics[height=.8in]{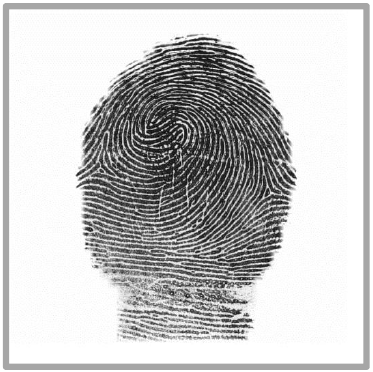}}~\raisebox{-.5\height}{\includegraphics[height=.8in]{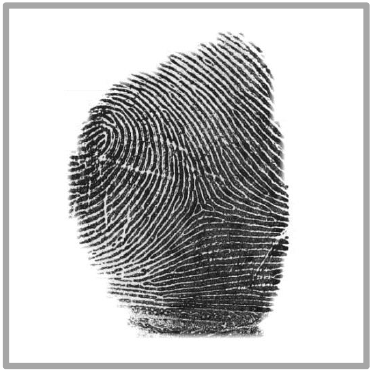}}
				& \raisebox{-.5\height}{\includegraphics[height=.8in]{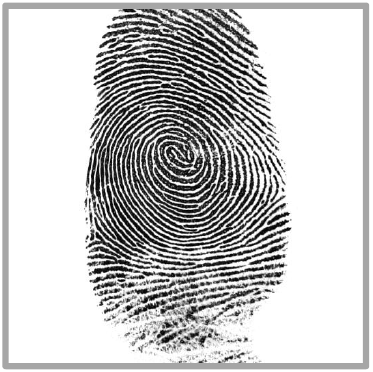}}
				&
				\raisebox{-.5\height}{\includegraphics[height=.8in]{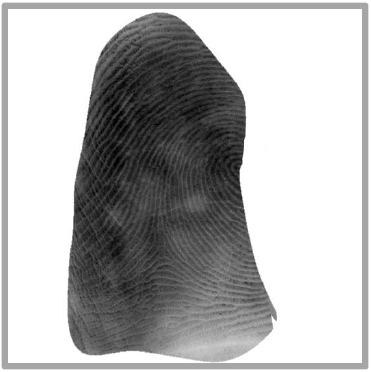}}~ \raisebox{-.5\height}{\includegraphics[height=.8in]{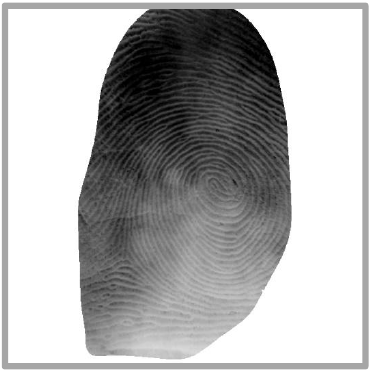}}~\raisebox{-.5\height}{\includegraphics[height=.8in]{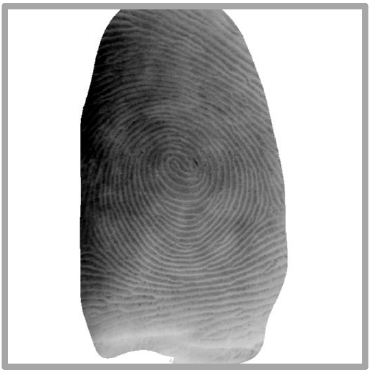}} \\
				\hhline
				Sensor                    
				& Optical
				& Optical
				& Camera  \\
				\hhline
				Description               
				& \multicolumn{1}{c}{\makecell{100 fingers $\times$ 3 poses \\ (left / middle / right)}}
				& \multicolumn{1}{c}{\makecell{200 fingers $\times$ 1}}
				& \multicolumn{1}{c}{\makecell{200 fingers $\times$ 3 poses\\ (left / middle / right) }} \\
				\hhline
				Usage                     
				& \multicolumn{1}{c}{\makecell{Registration accuracy}}
				& \multicolumn{2}{c}{\makecell{Registration accuracy\\Matching performance}} \\
				\hhline
				Genuine Match
				& 200\tnote{d}
				& \multicolumn{2}{c}{400 (CL-CL)\tnote{d} \ ,\quad 600 (C-CL)\tnote{e}}\\
				\hhline
				Imposter Match
				& $\backslash$
				& \multicolumn{2}{c}{19,900 (CL-CL)\tnote{f} \ ,\quad 19,900 (C-CL)\tnote{f}}\\
				\bottomrule
			\end{tabular}
			
			\begin{tablenotes}
				\item[a] Each fingerprint is matched with other fingerprints from the same finger. The symmetric matches are avoided.
				\item[b] First fingerprints of each finger are matched each other. The symmetric matches are avoided.
				\item[c] Each latent fingerprint is matched with all rolled fingerprints from other fingers.
				\item[d] All middle fingerprints are matched with the corresponding fingerprints on left and right side poses. The symmetric matches are avoided.
				\item[e] Each contact-based fingerprint is matched with other contactless fingerprints from the same finger. The symmetric matches are avoided.
				\item[f] A contactless fingerprint with specified pose in each group is sequentially selected and matched with other corresponding fingerprints. The symmetric matches are avoided.
						
			\end{tablenotes}
		\end{threeparttable}
	\end{center}
\end{table*}

\section{Dataset description}\label{sec:dataset}
Extensive experiments are conducted on multiple databases containing fingerprints with different impressions, including wildly used public datasets \emph{FVC2004 DB1\_A \& DB3\_A} \cite{fvc2004}, \emph{NIST SD27} \cite{nist27}, \emph{Tsinghua Distorted Fingerprint Database (TDF)}  \cite{si2015detection} and several private datasets \emph{Hisign Latent}, \emph{THU Old}, \emph{Hisign Multi-pose Plain Fingerprint Database (Hisign MPF)}, \emph{Hisign Contact-based 2D to Multi-pose Contactless 2D Fingerprint Database (Hisign C2CL)}.
Table \ref{tab:datasets} provides a comprehensive description about the composition and usage of these datasets.
In the following we will introduce the details of training and test data used in experiments.

\subsection{Training Data Building}
Similar to previous deep learning methods \cite{cui2019dense},\cite{cui2021dense}, we extract the real distortion field from \emph{TDF} and use it to transform fingerprints in \emph{Hisign Latent}, which enables us to obtain a large number of fingerprint pairs in two impressions and the ground truth of corresponding displacements.
Specifically, we use VeriFinger \cite{VeriFinger} to extract and track minutiae of distorted fingerprint videos in \emph{TDF}, and calculate TPS transform as the distortion field through paired points of first and last frames.
For a certain fingerprint $I$ in \emph{Hisign Latent}, a deformation field $D$ in \emph{TDF} is randomly selected and used for synthesis as:
\begin{equation}
	\begin{aligned}
		I^{\prime}\left(x+D_x, y+D_y\right) & = I(x, y), \\
		F \left(x, y\right) & = D(x, y), \\
		F^{\prime} \left(x+D_x, y+D_y\right) & = -D(x, y),
	\end{aligned}
\end{equation}
where $I^{\prime}$ is the conjugate fingerprint, $x$ and $y$ represent the corresponding direction components, $F$ and $F^{\prime}$ represent the displacement field registered from $I$ to $I^{\prime}$ and the opposite scenario respectively.
In this way, a total of $20,918$ sets similar to $\left\{ I, I^{\prime}, F \right\}$ are conveniently generated and used for network training.
In order to increase the diversity of fingerprint poses and categories, data augmentation strategies are used in the training stage, which include mirror flipping, rotation (by $90$, $180$ or $270$ degree) and swapping (as $\left\{I^{\prime}, I, F^{\prime}\right\}$).

\subsection{Matching Protocols in Testing}
In order to balance the number of genuine and imposter matches, specific protocols are set for different databases, as shown in Table \ref{tab:datasets}.
There are also some other settings: 
(i) considering \emph{FVC2004 DB1\_A} contains some strongly distorted fingerprints that may affect the accuracy of coarse alignment, we implemented a fingerprint distortion rectification process \cite{guan2023regression} marked with * in experiments to distinguish from the original dataset.
(ii) we do not evaluate the matching performance in \emph{Hisign MPF}, because the genuine and impostor scores of original images are already perfectly separated;
(iii) considering efficiency or fingerprint quality, subsets of \emph{Hisign C2CL} and \emph{THU Old} are selected and used in experiments;
(iv) in all datasets containing three poses, the genuine matching between the left and right side poses is not calculated because their overlapping area is too small;
(v) contactless-contactless (CL-CL) and contact-contactless (C-CL) matching experiments are implemented in \emph{Hisign C2CL} to evaluate the performance of fingerprint registration on multi-modality.

\section{Experiments}\label{sec:experiment}
In this section, we compare the proposed method with state-of-the-art algorithms.
TPS transformation based on matching minutiae is used as a benchmark for comparison due to its simplicity and practicality.
In addition, we compared phase registration \cite{cui2018phase}, which is representative of traditional methods, and two typical deep learning methods that estimate displacement locally \cite{cui2019dense} (called DRN (local)) or globally \cite{cui2021dense} (called DRN (global)).
The performance of fingerprint dense registration schemes are comprehensively evaluated in terms of registration accuracy, matching performance, and efficiency.
Moreover, ablation experiments are conducted to demonstrate the effectiveness of modules and strategies in PDRNet.

\begin{table*}[!t]
	\caption{Registration Accuracy of Different Fingerprint Registration Algorithms}
	\label{tab:registration}
	\vspace{-0.4cm}
	\begin{center}
		\begin{threeparttable}
			\begin{tabular}{p{.13\linewidth}<{\raggedright}*{12}{p{.042\linewidth}<{\centering}}}
				\toprule
				\multirow{2}*[-3pt]{\textbf{Method}}            
				& \multicolumn{2}{c}{\scriptsize\textbf{FVC2004 DB1\_A}}         
				& \multicolumn{2}{c}{\scriptsize\textbf{FVC2004 DB1\_A*}}         
				& \multicolumn{2}{c}{\scriptsize\textbf{Hisign MPF} }        
				& \multicolumn{2}{c}{\scriptsize\textbf{THU Old}}       
				& \multicolumn{2}{c}{\scriptsize\textbf{Hisign C2CL(CL-CL)}}                  
				& \multicolumn{2}{c}{\scriptsize\textbf{Hisign C2CL(C-CL)}} \\
				\cmidrule(lr){2-3}\cmidrule(lr){4-5}\cmidrule(lr){6-7}\cmidrule(lr){8-9}\cmidrule(lr){10-11}\cmidrule(lr){12-13}
				& NCC & VF
				& NCC & VF
				& NCC & VF
				& NCC & VF
				& NCC & VF
				& NCC & VF \\
				\midrule
				\multirow{1}{*}{TPS Based} 
				& 0.20 & 366
				& 0.19 & 362
				& 0.22 & 320
				& 0.20 & 322
				& 0.15 & 361
				& 0.09 & 321 \\
				\multirow{1}{*}{Phase Based\;\cite{cui2018phase}}
				& \textbf{0.65} & 404
				& \textbf{0.67} & 400
				& 0.62 & 350
				& 0.63 & 350
				& \textbf{0.60} & 391
				& \textbf{0.56} & 354 \\
				\multirow{1}{*}{DRN (local)\tnote{\dag}\;\cite{cui2019dense}}
				& 0.52 & 293
				& 0.54 & 289
				& 0.51 & 267
				& 0.52 & 269
				& 0.46 & 266
				& 0.43 & 221 \\
				\multirow{1}{*}{DRN (global)\tnote{\dag}\;\cite{cui2021dense}}
				& 0.55 & 388
				& 0.58 & 388
				& 0.56 & 340
				& 0.56 & 343
				& 0.51 & 375
				& 0.45 & 335 \\
				\midrule
				\multirow{1}{*}{Proposed\tnote{\dag}}
				& \textbf{0.65} & \textbf{417}
				& \textbf{0.67} & \textbf{410}
				& \textbf{0.64} & \textbf{357}
				& \textbf{0.65} & \textbf{353}
				& 0.59 & \textbf{397}
				& \textbf{0.56} & \textbf{368} \\
				\bottomrule
			\end{tabular}
			\begin{tablenotes}
				\item[\dag] Registration method based on neural networks.
			\end{tablenotes}
		\end{threeparttable}
	\end{center}
\end{table*}

\begin{figure*}[!t]
	\centering
	\includegraphics[width=1\linewidth]{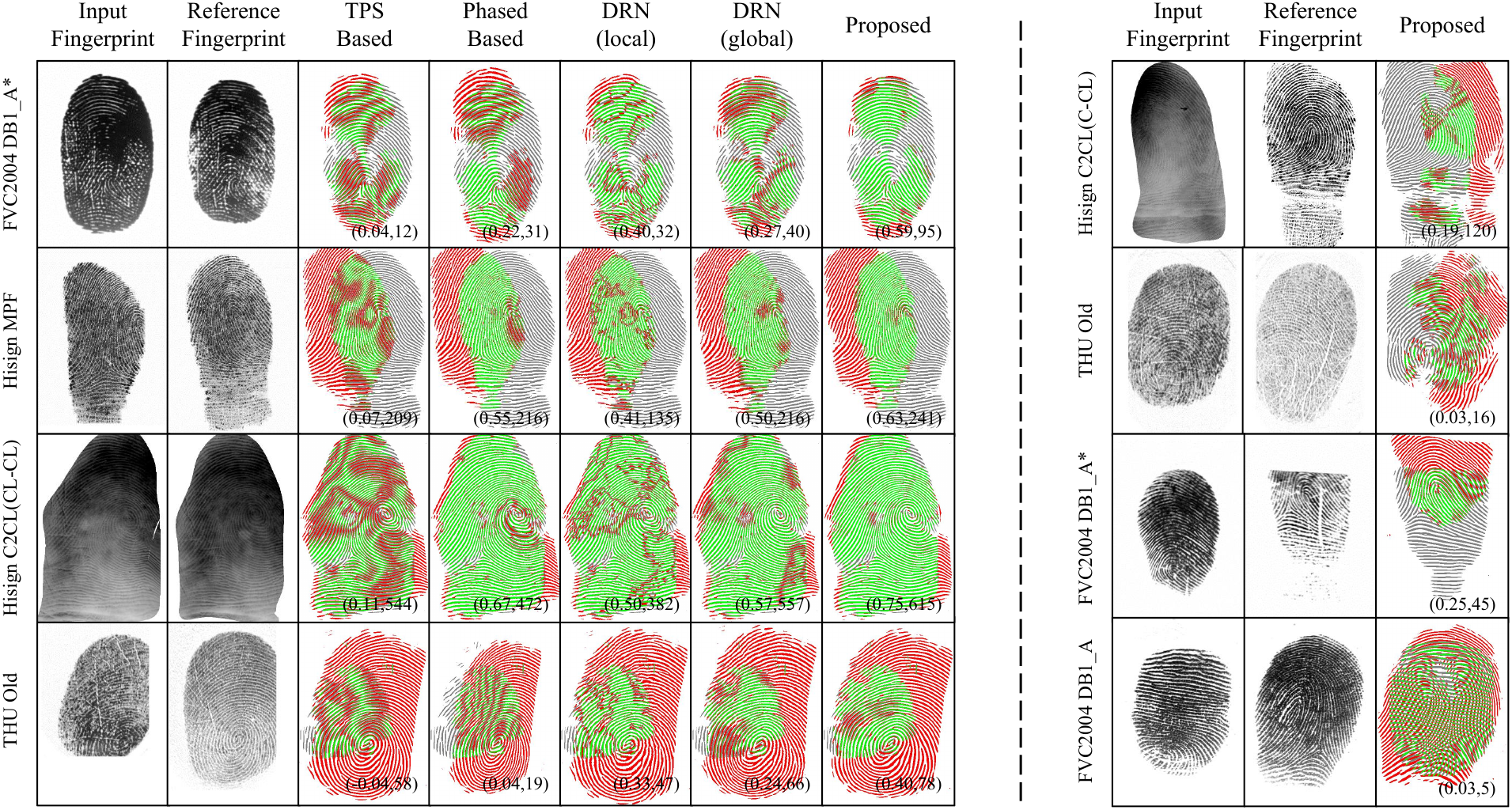}
	\caption{Examples of fingerprint registration for genuine matching fingerprints. 
		The left side compares the performance of different registration methods in typical scenarios. 
		Some representative cases where our approach fails are shown on the right. 
		The beginning of each row gives the name of corresponding datasets, and numbers in brackets are matching
		scores by image correlator and VeriFinger. 
		Green indicates overlap, while red and gray indicate non-overlapping areas of respective fingerprints.}
	\label{fig:registration_example}
\end{figure*}

\begin{figure}[!t]
	\centering
	\includegraphics[width=.95\linewidth]{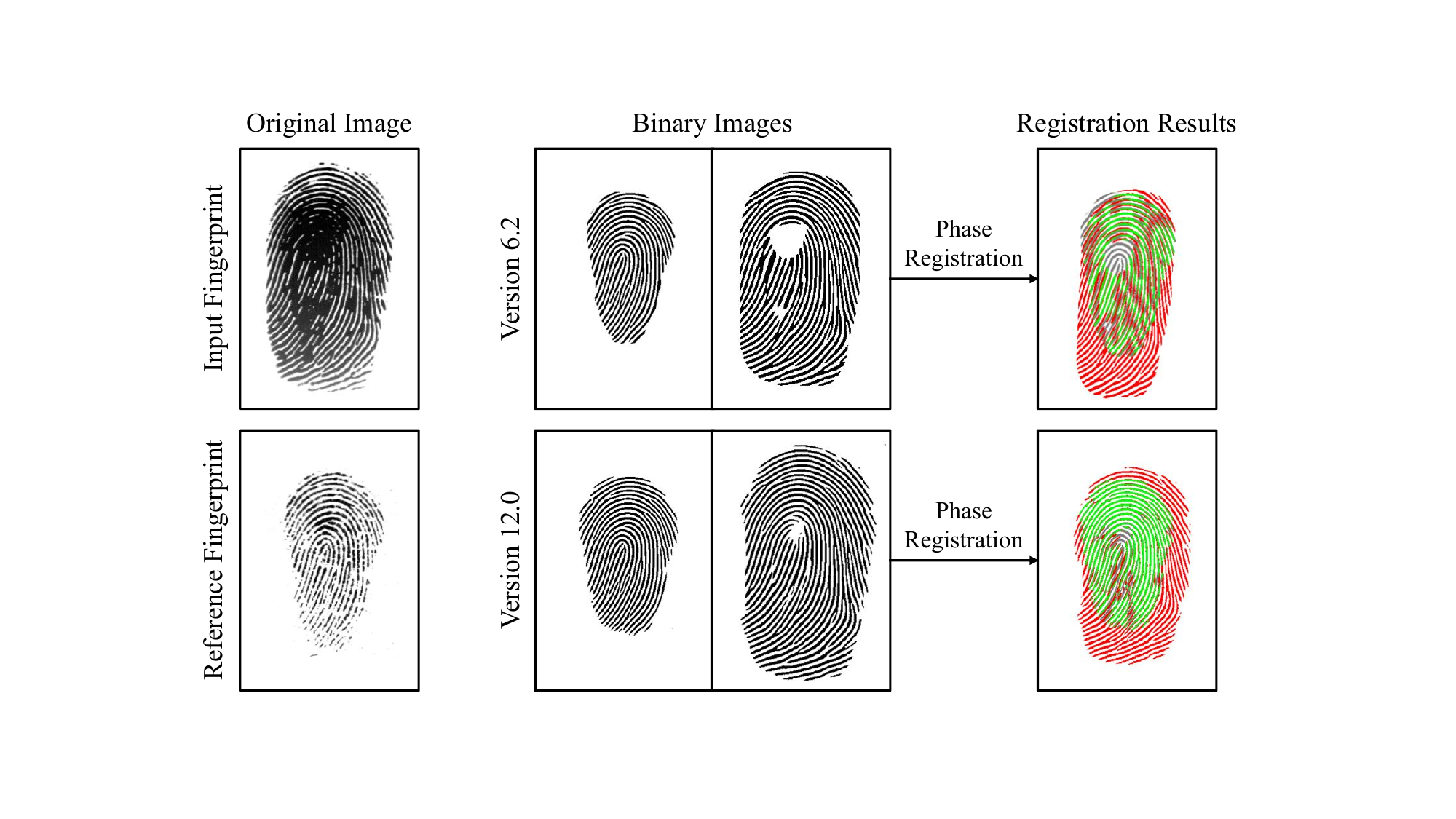}
	\caption{Registration results based on phase information with binary images extracted by different versions of VeriFinger. Green indicates overlap of two fingerprint ridges, while gray and red indicate no overlap.}
	\label{fig:phase_version}
\end{figure}

\subsection{Evaluation Protocols}

Image correlator and VeriFinger \cite{VeriFinger} are used to reflect the similarity between two fingerprints, consistent with previous works \cite{si2017dense},\cite{cui2018phase},\cite{cui2019dense},\cite{lan2020preregistration},\cite{cui2021dense}.
Let $I$ and $M$ represent the image and mask of fingerprints, for any two fingerprints $1$ and $2$ the correlation score is calculated as
\begin{equation}
	\begin{aligned}
		NCC & =  \frac{\sum_{M}\left(I_1-\overline{I_1}\right)\cdot\left(I_2-\overline{I_2}\right)}{\sqrt{\sum_{M}\left(I_1-\overline{I_1}\right)^2\cdot\sum_{M}\left(I_2-\overline{I_2}\right)^2}},
	\end{aligned}
	\label{eq:ncc}
\end{equation}
where $M$ is the common area of $M_1$ and $M_2$, $\overline{I}$ represents the mean value of image $I$ in $M$.
This metric can sensitively reflect the degree of ridge overlapping and is easy to implement. 
Since the experiments mainly focus on the relative improvement brought by registration algorithms rather than the absolute performance, the image correlator can be regarded as a representative of image-based matchers.
On the other hand, we choose to employ VeriFinger SDK 12.0 \cite{VeriFinger}, a widely used commercial software, to measure the alignment of minutiae due to its superior performance in this type of matchers.
For convenience, the matching score of VeriFinger is referred to as \textit{VF} in the following.
%{\color{blue} It should be noted that matching methods based on global descriptors, such as \cite{engelsma2021learning,grosz2022c2cl}, are not applied in this paper because they may not pay equal attention to every local fine structures, that is to say, the scores obtained in this manner lack sufficient interpretability for the performance of dense fingerprint registration. Furthermore, fixed-length global descriptors have limited capability in distinguishing relative translation differences of ridges compared to image correlation, which is precisely concerned in dense fingerprint registration.}
It should be noted that matching methods based on global descriptors, such as \cite{engelsma2021learning,grosz2022c2cl}, are not applied in this paper because they may not pay equal attention to every local fine structures, that is to say, the scores obtained in this manner lack sufficient interpretability for the performance of dense fingerprint registration.
Furthermore, fixed-length global descriptors have limited capability in distinguishing relative translation differences of ridges compared to image correlation, which is precisely concerned in dense fingerprint registration.

\subsection{Registration Accuracy} \label{subsec:registration_accuracy}

\begin{table*}[!t]
	\caption{Matching Performance by Image Correlator with Different Fingerprint Registration Algorithms}
	\label{tab:matching_corr}
	\vspace{-0.4cm}
	\begin{center}
		\begin{threeparttable}
			\begin{tabular}{p{.14\linewidth}<{\raggedright}*{18}{p{.022\linewidth}<{\centering}}}
				\toprule
				\multirow{2}*[-3pt]{\textbf{Method}}            
				& \multicolumn{3}{c}{\scriptsize\textbf{FVC2004 DB1\_A}}         
				& \multicolumn{3}{c}{\scriptsize\textbf{FVC2004 DB1\_A*}}         
				& \multicolumn{3}{c}{\scriptsize\textbf{FVC2004 DB3\_A} }        
				& \multicolumn{3}{c}{\scriptsize\textbf{THU Old}}       
				& \multicolumn{3}{c}{\scriptsize\textbf{Hisign C2CL(CL-CL)}}                  
				& \multicolumn{3}{c}{\scriptsize\textbf{Hisign C2CL(C-CL)}} \\
				\cmidrule(lr){2-4}\cmidrule(lr){5-7}\cmidrule(lr){8-10}\cmidrule(lr){11-13}\cmidrule(lr){14-16}\cmidrule(lr){17-19}
				& \scriptsize{TAR} & \scriptsize{FMR} & \scriptsize{EER}
				& \scriptsize{TAR} & \scriptsize{FMR} & \scriptsize{EER}
				& \scriptsize{TAR} & \scriptsize{FMR} & \scriptsize{EER}
				& \scriptsize{TAR} & \scriptsize{FMR} & \scriptsize{EER}
				& \scriptsize{TAR} & \scriptsize{FMR} & \scriptsize{EER}
				& \scriptsize{TAR} & \scriptsize{FMR} & \scriptsize{EER} \\
				\midrule
				\multirow{1}{*}{TPS Based} 
				& 70.2 & 33.8 & 11.5
				& 72.6 & 46.9 & 11.6
				& 74.6 & 65.4 & 10.4
				& 70.6 & 37.2 & 11.7
				& 61.4 & 78.1 & 12.5
				& 45.3 & 99.6 & 22.3 \\
				\multirow{1}{*}{Phase Based\;\cite{cui2018phase}}
				& 97.8 & 3.16 & 1.12
				& 98.7 & 2.78 & 0.64
				& 93.1 & 9.23 & 1.79
				& 94.9 & 9.53 & 3.68
				& 97.8 & \textbf{13.0} & 0.96
				& 96.2 & 48.3 & 1.17\\
				\multirow{1}{*}{DRN (local)\tnote{\dag}\;\cite{cui2019dense}}
				& 91.1 & 25.5 & 2.49
				& 87.2 & 20.8 & 2.75
				& 72.5 & 46.4 & 5.98
				& 76.5 & 53.5 & 8.65
				& 71.8 & 82.7 & 5.75
				& 67.0 & 92.3 & 6.67\\
				\multirow{1}{*}{DRN (global)\tnote{\dag}\;\cite{cui2021dense}}
				& 95.4 & 12.1 & 2.42
				& 97.5 & 8.00 & 1.10
				& 91.8 & 30.1 & 2.01
				& 91.9 & 14.7 & 2.21
				& 95.8 & 28.2 & 1.25
				& 95.3 & 42.8 & 1.19\\
				\midrule
				\multirow{1}{*}{Proposed\tnote{\dag}}
				& \textbf{98.9} & \textbf{1.54} & \textbf{0.79}
				& \textbf{99.8} & \textbf{0.41} & \textbf{0.17}
				& \textbf{98.4} & \textbf{2.78} & \textbf{1.24}
				& \textbf{98.5} & \textbf{2.21} & \textbf{1.47}
				& \textbf{99.5} & 21.7 & \textbf{0.25}
				& \textbf{99.5} & \textbf{4.16} & \textbf{0.33}\\
				
				\bottomrule
			\end{tabular}
			\begin{tablenotes}
				\item[\dag] Registration method based on neural networks.
				\item[] TAR represents TAR@FAR = 0.1\%, FMR represents ZeroFMR.
			\end{tablenotes}
		\end{threeparttable}
	\end{center}
\end{table*}

\begin{table*}[!t]
	\caption{Matching Performance by VeriFinger Matcher with Different Fingerprint Registration Algorithms}
	\label{tab:matching_verifinger}
	\vspace{-0.4cm}
	\begin{center}
		\begin{threeparttable}
			\begin{tabular}{p{.14\linewidth}<{\raggedright}*{18}{p{.022\linewidth}<{\centering}}}
				\toprule
				\multirow{2}*[-3pt]{\textbf{Method}}            
				& \multicolumn{3}{c}{\scriptsize\textbf{FVC2004 DB1\_A}}         
				& \multicolumn{3}{c}{\scriptsize\textbf{FVC2004 DB1\_A*}}         
				& \multicolumn{3}{c}{\scriptsize\textbf{FVC2004 DB3\_A} }        
				& \multicolumn{3}{c}{\scriptsize\textbf{THU Old}}       
				& \multicolumn{3}{c}{\scriptsize\textbf{Hisign C2CL(CL-CL)}}                  
				& \multicolumn{3}{c}{\scriptsize\textbf{Hisign C2CL(C-CL)}} \\
				\cmidrule(lr){2-4}\cmidrule(lr){5-7}\cmidrule(lr){8-10}\cmidrule(lr){11-13}\cmidrule(lr){14-16}\cmidrule(lr){17-19}
				& \scriptsize{TAR} & \scriptsize{FMR} & \scriptsize{EER}
				& \scriptsize{TAR} & \scriptsize{FMR} & \scriptsize{EER}
				& \scriptsize{TAR} & \scriptsize{FMR} & \scriptsize{EER}
				& \scriptsize{TAR} & \scriptsize{FMR} & \scriptsize{EER}
				& \scriptsize{TAR} & \scriptsize{FMR} & \scriptsize{EER}
				& \scriptsize{TAR} & \scriptsize{FMR} & \scriptsize{EER} \\
				\midrule
				\multirow{1}{*}{TPS Based} 
				& 98.8 & 1.86 & 0.75
				& 99.4 & 1.61 & 0.45
				& 98.9 & \textbf{1.32} & 0.70
				& 96.3 & 5.15 & 1.60
				& 99.7 & 0.69 & 0.26
				& 99.7 & \textbf{0.33} & 0.32 \\
				\multirow{1}{*}{Phase Based\;\cite{cui2018phase}}
				& \textbf{99.2} & \textbf{1.04} & 0.61
				& \textbf{99.5} & \textbf{0.57} & 0.41
				& 98.6 & 1.81 & 0.80
				& 97.1 & 3.41 & 2.20
				& 99.7 & \textbf{0.25} & 0.25
				& 99.7 & 0.91 & 0.32 \\
				\multirow{1}{*}{DRN (local)\tnote{\dag}\;\cite{cui2019dense}}
				& 97.2 & 5.32 & 1.62
				& 97.7 & 2.68 & 1.28
				& 95.9 & 5.64 & 2.13
				& 93.4 & 8.09 & 4.21
				& 98.3 & 5.25 & 0.78
				& 98.7 & 4.33 & 1.20\\
				\multirow{1}{*}{DRN (global)\tnote{\dag}\;\cite{cui2021dense}}
				& 98.6 & 2.37 & 0.83
				& 99.4 & 1.14 & 0.36
				& 98.8 & 1.71 & 0.72
				& 97.8 & 4.32 & 2.25
				& 99.5 & 1.00 & 0.49
				& 99.7 & 0.83 & 0.32 \\
				\midrule
				\multirow{1}{*}{Proposed\tnote{\dag}}
				& \textbf{99.2} & 1.16 & \textbf{0.58}
				& \textbf{99.5} & 0.71 & \textbf{0.23}
				& \textbf{99.1} & 1.52 & \textbf{0.66}
				& \textbf{98.5} & \textbf{1.47} & \textbf{1.41}
				& \textbf{100} & 0.50 & \textbf{0.02}
				& \textbf{99.8} & 0.58 & \textbf{0.16} \\
				\bottomrule
			\end{tabular}
			\begin{tablenotes}
				\item[\dag] Registration method based on neural networks.
				\item[] TAR represents TAR@FAR = 0.1\%, FMR represents ZeroFMR.
			\end{tablenotes}
		\end{threeparttable}
	\end{center}
\end{table*}

\begin{figure}[!t]
	\centering
	{\includegraphics[width=1\linewidth]{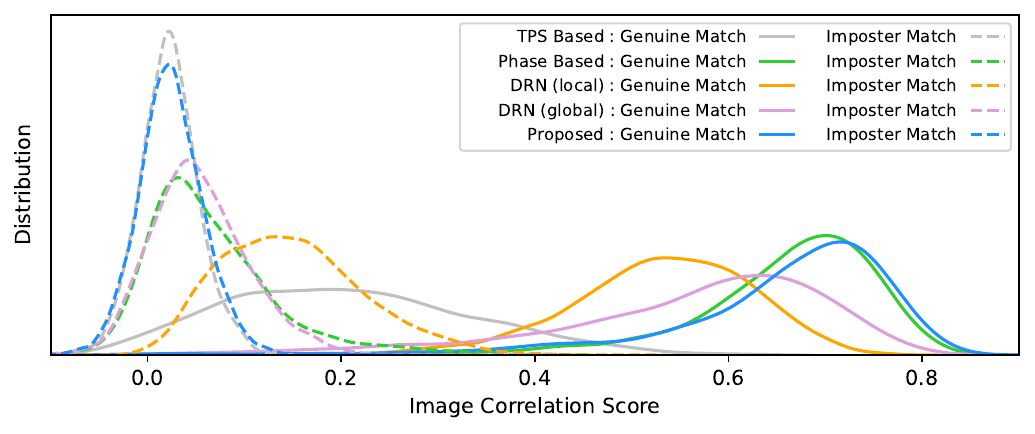}%
	}
	
	{\includegraphics[width=1\linewidth]{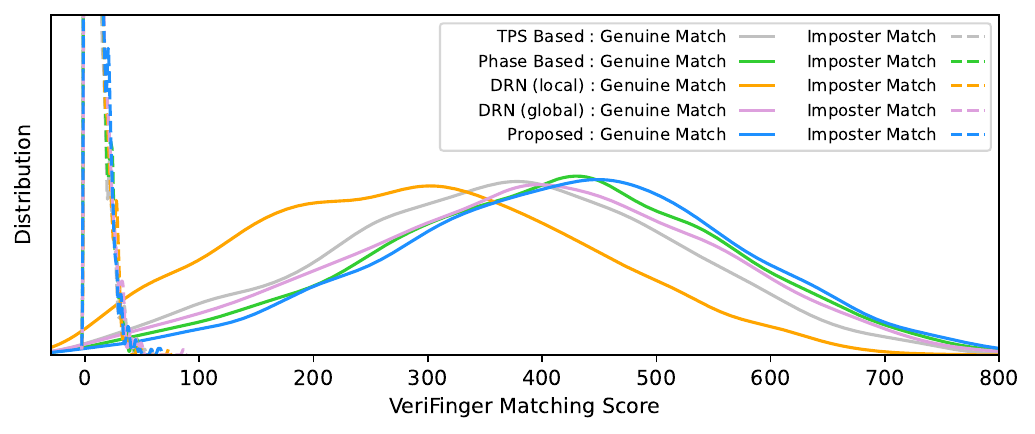}%
	}
	\caption{Distribution of genuine and imposter matching scores on FVC2004 DB1\_A by image correlator (top) and VeriFinger matcher (bottom). The scale on vertical axis is not displayed because we are more concerned with the relative values of probability density.}
	\label{fig:distribution}
\end{figure}

\begin{figure*}[!t]
	\centering
	\subfloat[FVC2004 DB1\_A]{\includegraphics[width=.33\linewidth]{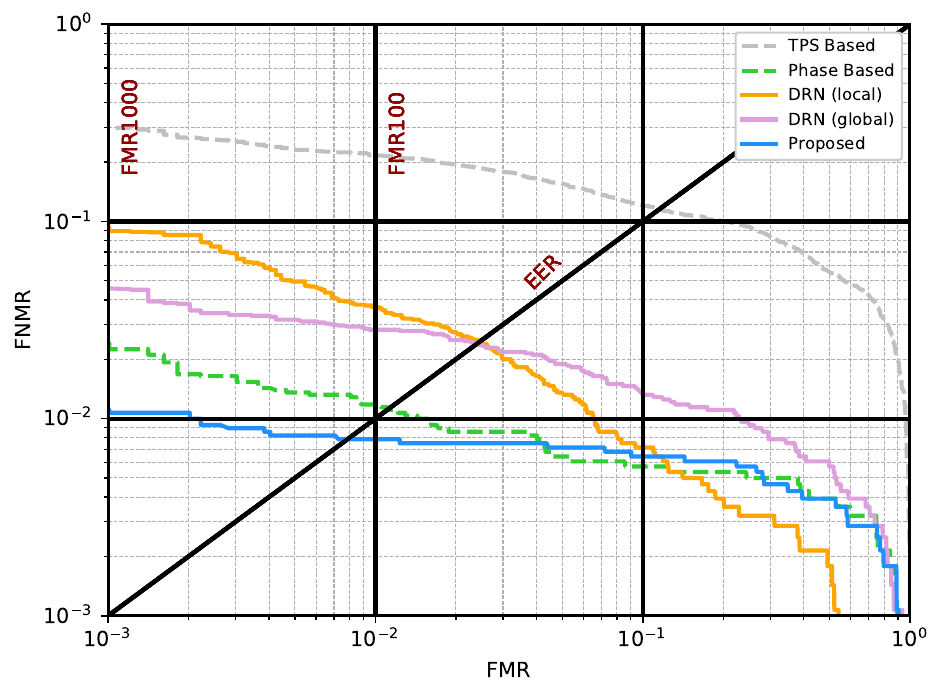}%
	}
	\hfil
	\subfloat[FVC2004 DB1\_A*]{\includegraphics[width=.33\linewidth]{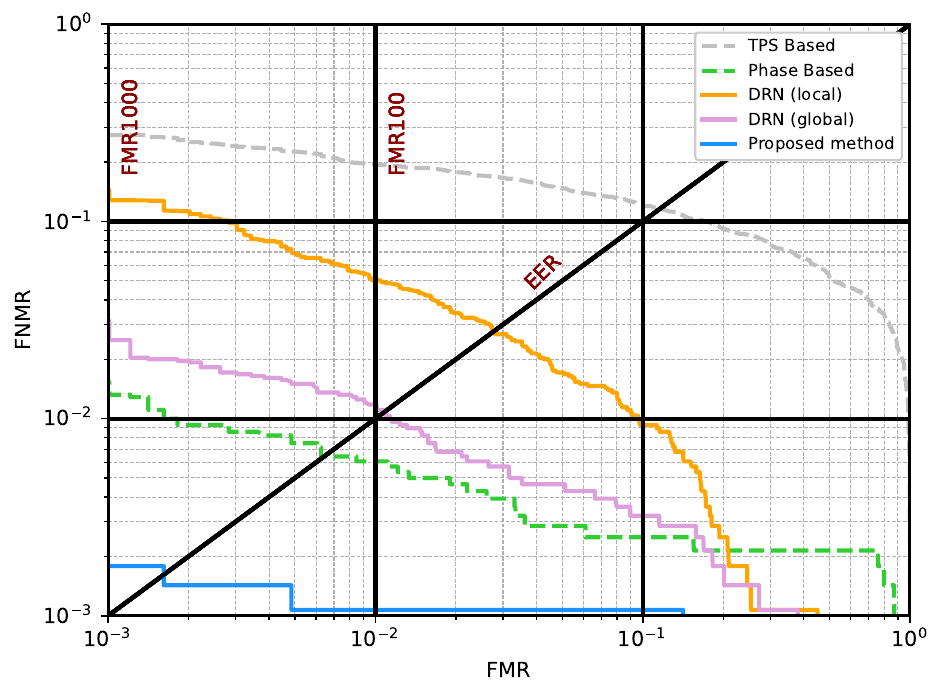}%
	}
	\hfil
	\subfloat[FVC2004 DB3\_A]{\includegraphics[width=.33\linewidth]{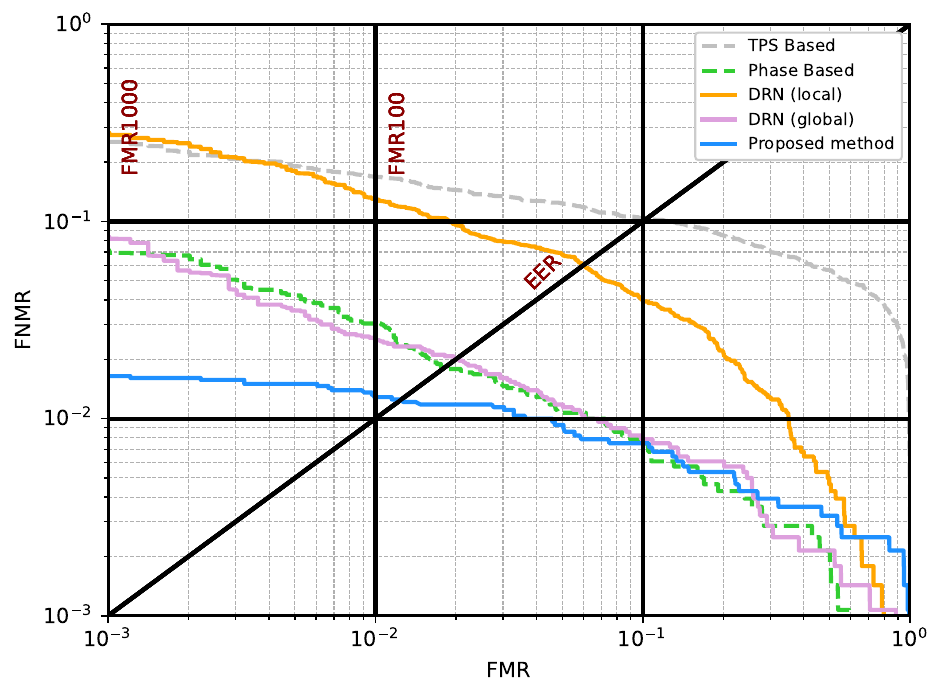}%
	}
	
	\subfloat[THU Old]{\includegraphics[width=.33\linewidth]{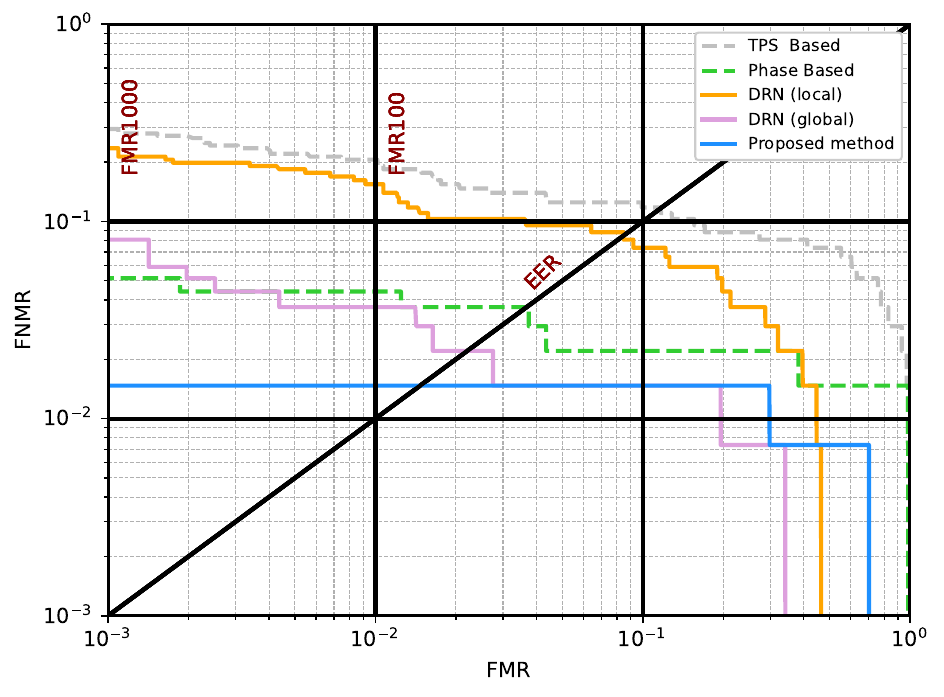}%
	}
	\hfil
	\subfloat[Hisign C2CL(CL-CL)]{\includegraphics[width=.33\linewidth]{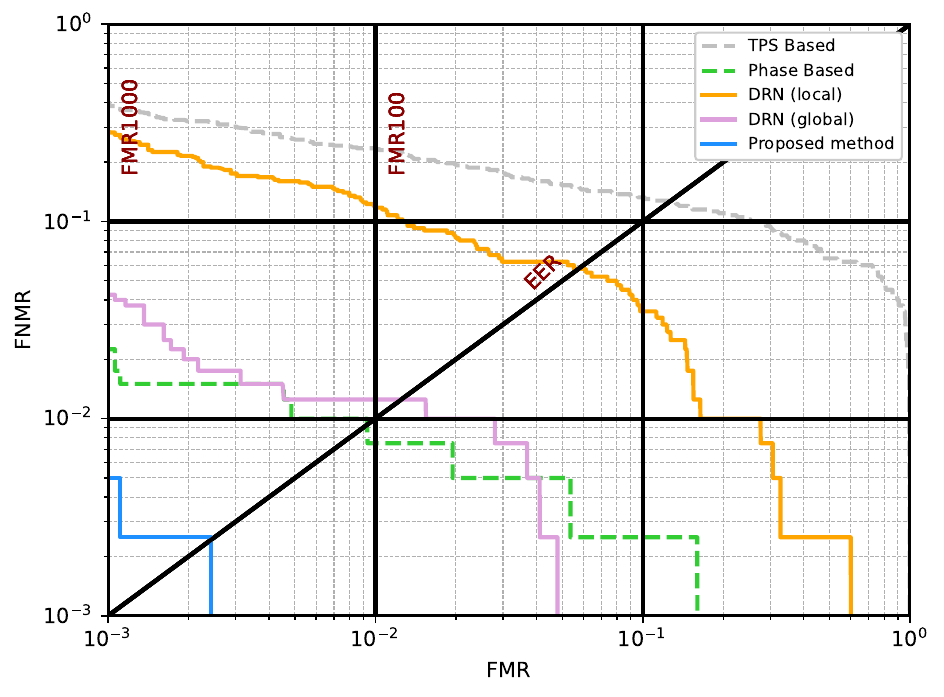}%
	}
	\hfil
	\subfloat[Hisign C2CL(C-CL)]{\includegraphics[width=.33\linewidth]{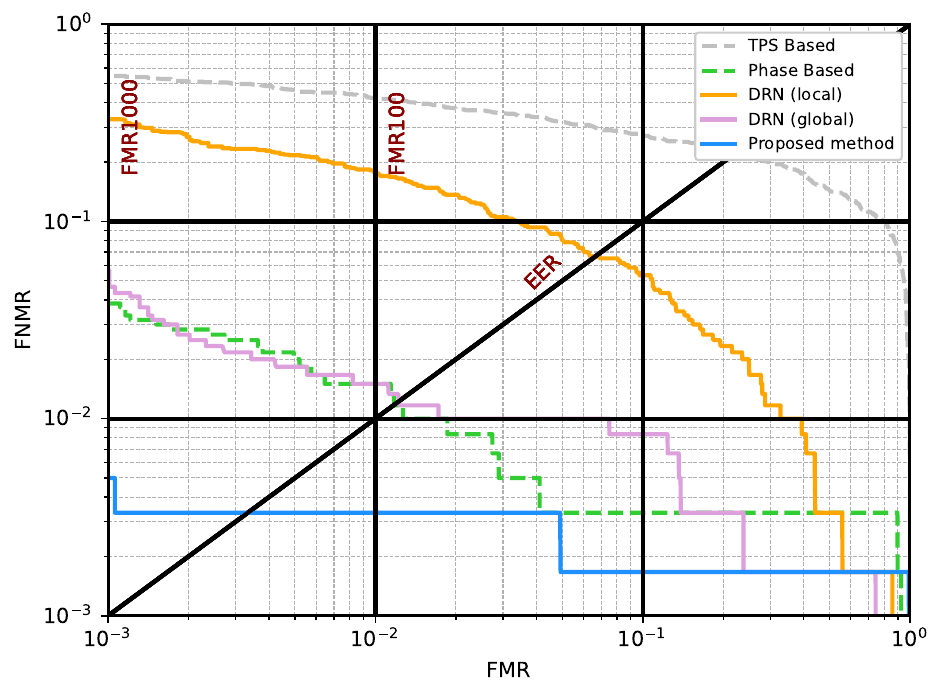}%
	}
	\caption{DET curves by image correlator. Solid and dotted lines represent deep learning methods and traditional methods respectively.}
	\label{fig:det_corr}
\end{figure*}

\begin{figure*}[!t]
	\centering
	\subfloat[FVC2004 DB1\_A]{\includegraphics[width=.33\linewidth]{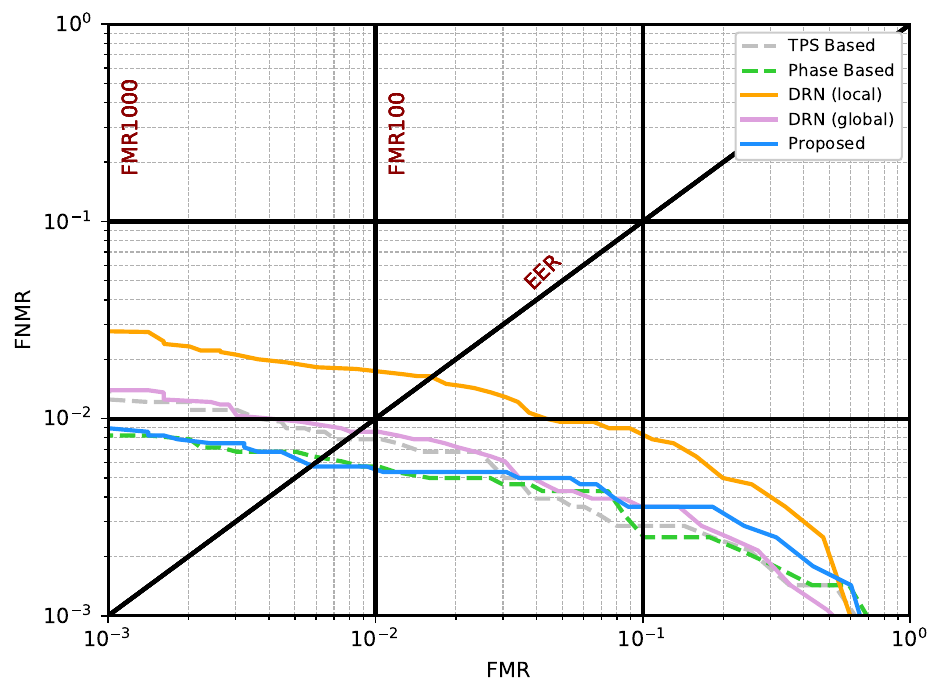}%
	}
	\hfil
	\subfloat[FVC2004 DB1\_A*]{\includegraphics[width=.33\linewidth]{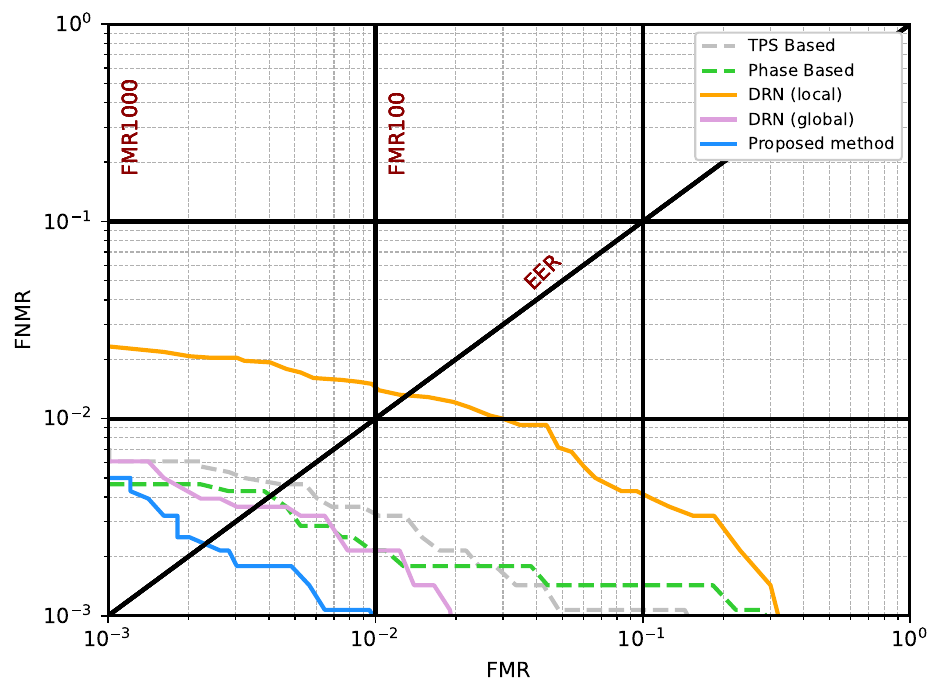}%
	}
	\hfil
	\subfloat[FVC2004 DB3\_A]{\includegraphics[width=.33\linewidth]{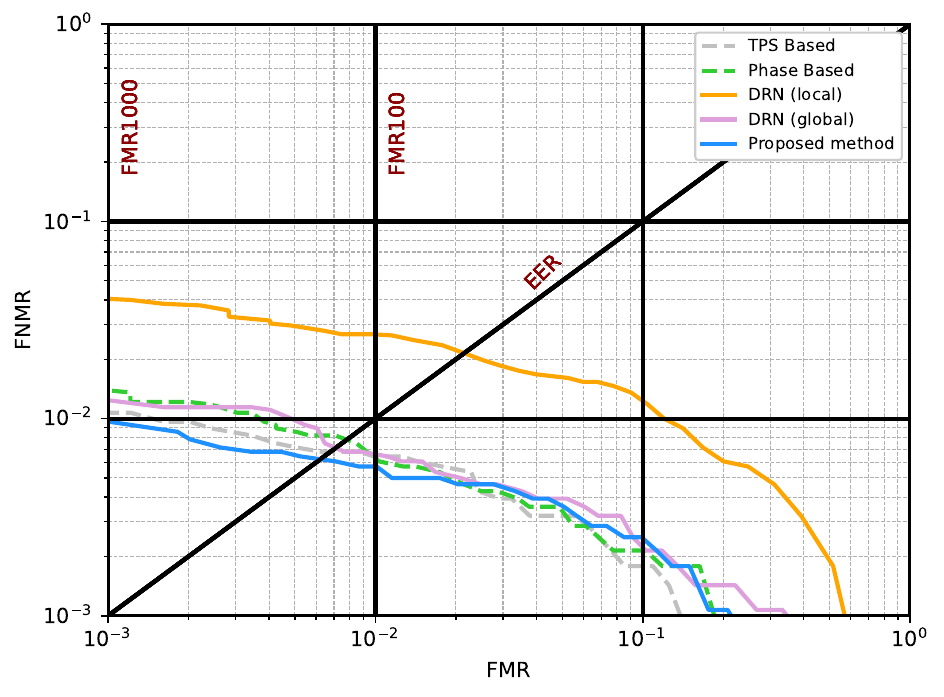}%
	}
	
	\subfloat[THU Old]{\includegraphics[width=.33\linewidth]{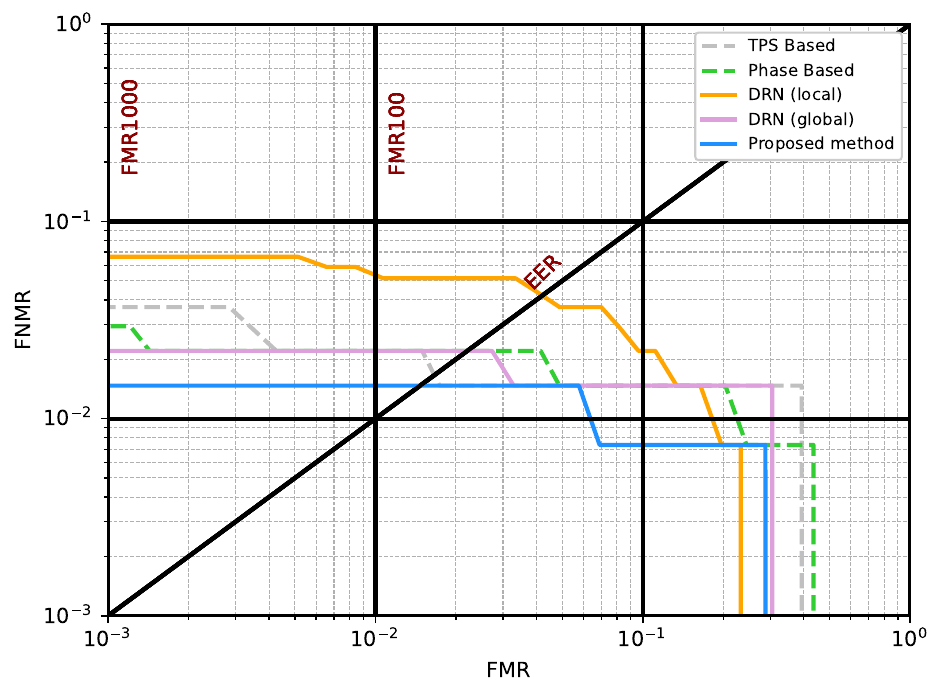}%
	}
	\hfil
	\subfloat[Hisign C2CL(CL-CL)]{\includegraphics[width=.33\linewidth]{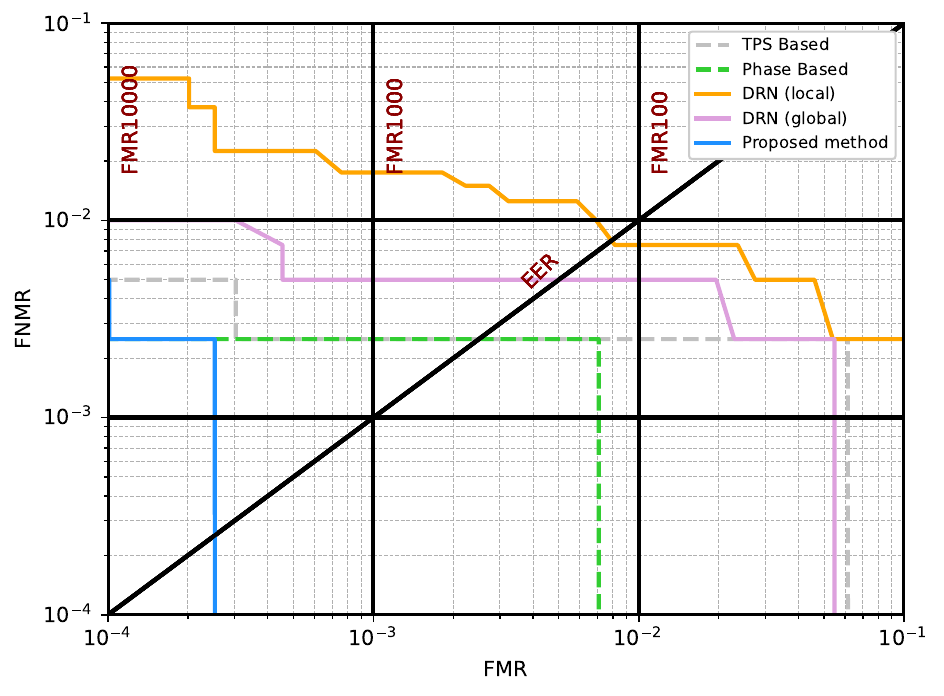}%
	}
	\hfil
	\subfloat[Hisign C2CL(C-CL)]{\includegraphics[width=.33\linewidth]{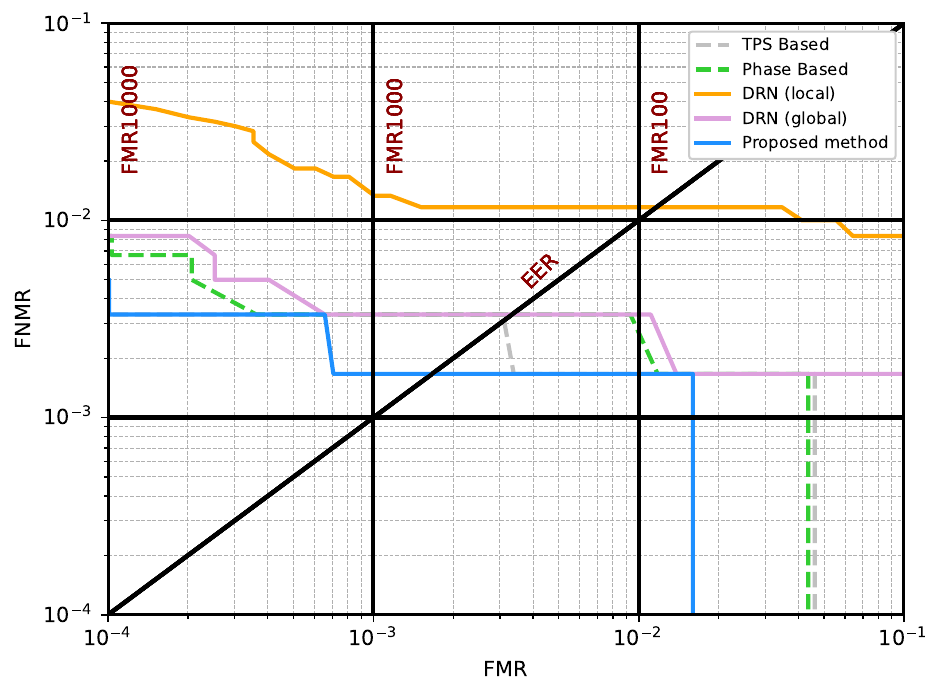}%
	}
	\caption{DET curves by VeriFinger matcher. Solid and dotted lines represent deep learning methods and traditional methods respectively.}
	\label{fig:det_verifinger}
\end{figure*}

In this subsection, genuine matching are conducted using image correlator and VeriFinger matcher to quantitatively evaluate the contribution of dense registration algorithms to the alignment of ridges and minutiae. 
We utilize databases \emph{FVC2004 DB1\_A}, \emph{Hisign MPF}, \emph{THU Old} and \emph{Hisign C2CL} because they fit the scenarios in practical applications, such as fingerprint mosaicking and cross-modal comparison.

As shown in Table \ref{tab:registration}, the proposed method significantly outperforms other learning based methods in all datasets and maintains advantages over conventional methods.
This suggests that our network is capable of establishing dense relationships between fingerprint pairs more precisely while ensuring high stability of relative structures in space.

Several typical examples are given in Fig. \ref{fig:registration_example} in order to show the effect of fingerprint registration methods more intuitively. 
From the comparisons on the left, it can be seen that phase registration \cite{cui2018phase} is difficult to deal with scenes of incomplete (low-quality areas in row 1), complex texture structures (singular regions in rows 2 and 3) and interlaced ridges (folds and distortion in row 4).
On the other hand, the local displacement registration network \cite{cui2019dense} is more robust in the above areas, but it only gives a local optimal solution which is burr in space and obviously reflected in registration results.
Subsequent improvements proposed by Cui \etal \cite{cui2021dense} significantly improved the global smoothness, but there are still misalignments in some ridge areas. 
One reasonable explanation is that the network tends to learn features of point structures without additional guidance (see Fig. \ref{fig:heatmap}), which are just lacking in these problem regions.
Our method integrates the advantages of traditional algorithm and deep learning, showing higher accuracy and stability in registration.

%{\color{blue}Meanwhile, the representative failure cases on the right of Fig. \ref{fig:registration_example} show that our algorithm still needs to be improved in some extreme scenarios, such as: (i) image defects caused by low contrast or extensive wrinkles (row 1 and 2); (ii) lack of sufficient reference information due to limited overlap (row 3); (iii) severe spatial structural dislocation caused by large distortion (row 4).}
Meanwhile, the representative failure cases on the right of Fig. \ref{fig:registration_example} show that our algorithm still needs to be improved in some extreme scenarios, such as: (i) image defects caused by low contrast or extensive wrinkles (row 1 and 2); (ii) lack of sufficient reference information due to limited overlap (row 3); (iii) severe spatial structural dislocation caused by large distortion (row 4).

It should be mentioned that the performance of phase based registration \cite{cui2018phase} in this paper is higher than those reported in previous papers \cite{cui2018phase,cui2019dense,lan2020preregistration,cui2021dense}. This is because we use a newer version of VeriFinger (from 6.2 to 12.0) which performs better in extracting binary images. Fig. \ref{fig:phase_version} shows the impact of software version on phase registration.

\subsection{Matching Performance} \label{subsec:matching_performance}

Further experiments are conducted on several databases with multiple modalities, including three widely used public datasets \emph{FVC2004 DB1\_A}, \emph{FVC2004 DB3\_A}, \emph{NIST SD27} and two additional private datasets \emph{THU Old}, \emph{Hisign C2CL}, to examine the assistance of fingerprint registration algorithms on matching performance. Similar to previous works, we evaluate the similarity score of mated or non-mated fingerprint pairs using image correlator and VeriFinger matcher.

We first present the score distribution of genuine and imposter matches on \emph{FVC2004 DB1\_A} to qualitatively assess the performance of different registration algorithms.
As shown in Fig. \ref{fig:distribution}, our proposed method outperforms the others in improving genuine matching scores.
In addition, our method obviously has more concentrated distribution and lower average score on imposter matches of image correlation.
DRN (local) \cite{cui2019dense} exhibits the poorest performance because it lacks global constraints. 
Other suboptimal methods \cite{cui2018phase},\cite{cui2021dense} may align local blocks of non-mated fingerprints although global information exchange are introduced explicitly or implicitly.
As analyzed above, our network aggregates the advantages of traditional method and deep learning, which simultaneously focuses on the characteristics of ridge lines and anchor points while conducting global information interaction in multiple stages, thus achieving the highest precision and stability.

Table \ref{tab:matching_corr} and Table \ref{tab:matching_verifinger} show the comparison of matching performance after different registration methods.
Three representative indicators True Accept Rate (TAR), False Match Rate (FMR) and Equal Error Rate (EER), which are commonly used in biometric recognition systems, are listed to briefly reflect the accuracy in identification scenarios.
The corresponding Detection Error Tradeoff (DET) curves are shown in Fig. \ref{fig:det_corr} and Fig. \ref{fig:det_verifinger} for more complete information.
Experimental results demonstrate that the proposed method surpasses other methods in almost all evaluation aspects, and its advantages in TAR and EER are particularly obvious.

\begin{figure*}[!t]
	\centering
	\subfloat[Image correlator]{\includegraphics[width=.45\linewidth]{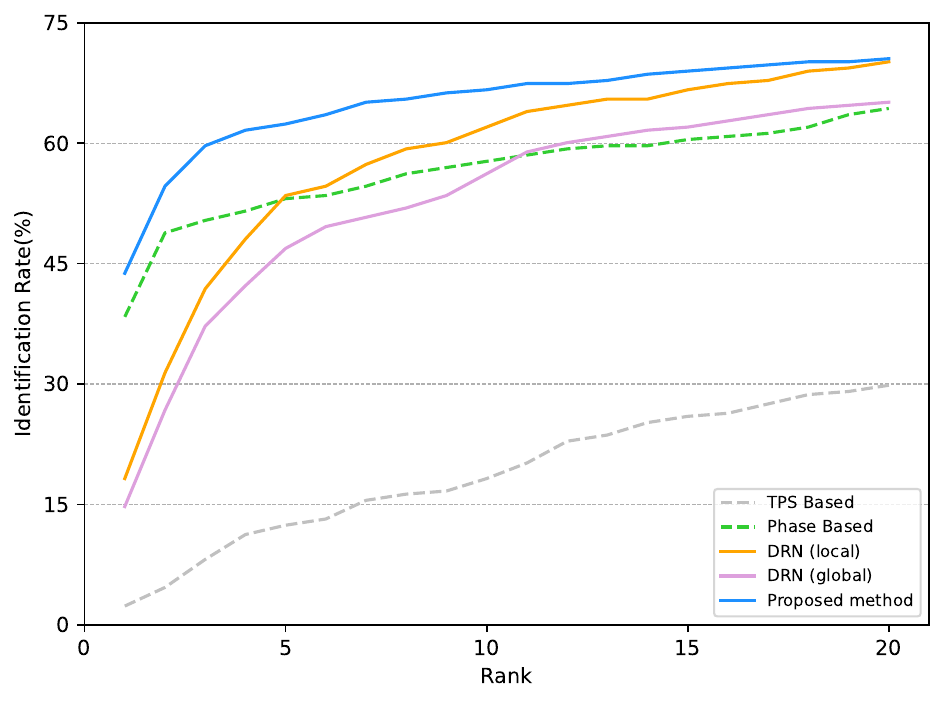}%
	}
	\hfil
	\subfloat[VeriFinger matcher]{\includegraphics[width=.45\linewidth]{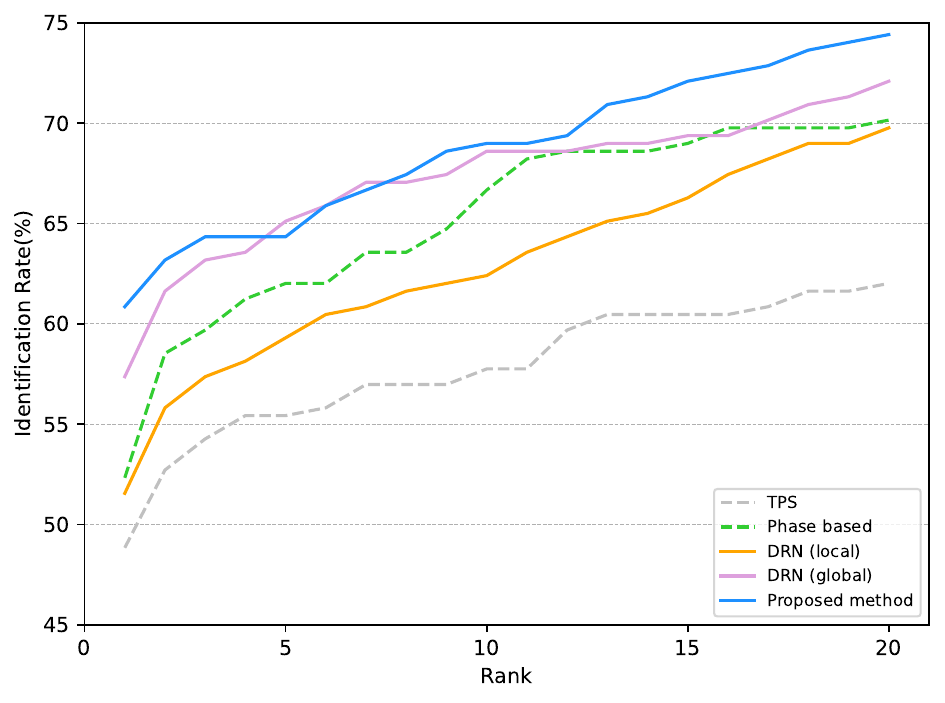}%
	}
	\caption{CMC curves by image correlator and VeriFinger matcher with different fingerprint registration algorithms on NIST SD27. Solid and dotted lines represent deep learning methods and traditional methods respectively.}
	\label{fig:cmc}
\end{figure*}

Furthermore, we calculate the Cumulative Matching Characteristic (CMC) curves on latent fingerprint database \emph{NIST SD27} to evaluate the performance of registration algorithms on low-quality fingerprints.
As shown in Fig. \ref{fig:cmc}, our method is most robust to these complex and difficult samples.
It is worth mentioning that deep learning based method \cite{cui2021dense} perform better than traditional method \cite{cui2018phase}, which is different from the results on other datasets.
This phenomenon occurs because the dense presence of incomplete or contaminated areas in latent fingerprints disrupts normal feature extraction, which usually leads to misjudgments by those algorithms.
In contrast, deep learning methods can reduce the interference of local misinformation by selectively using more stable features.

\begin{table}[!t]
	\caption{Ablation Study of the Proposed Network with Different Modules and Strategies on Hisign MPF}
	\label{tab:ablation_structure}
	\vspace{-0.4cm}
	\begin{center}
		\begin{threeparttable}
			\begin{tabular}{*{1}{p{0.14\linewidth}<{\centering}}*{1}{p{0.09\linewidth}<{\centering}}*{1}{p{0.14\linewidth}<{\centering}}*{2}{p{0.09\linewidth}<{\centering}}*{1}{p{.15\linewidth}<{\centering}}}
				\toprule
				\multicolumn{5}{c}{\scriptsize{\textbf{Modules \& Strategies}}}
				& \multirow{2}*[-7pt]{\scriptsize{\textbf{NCC}}} \\
				\cmidrule(lr){1-5}
				\multicolumn{1}{c}{\scriptsize{\makecell{Correlation\\branch}}}
				& \multicolumn{1}{c}{\scriptsize{\makecell{Context\\branch}}}
				& \multicolumn{1}{c}{\scriptsize{\makecell{Information\\interaction}}}
				& \multicolumn{1}{c}{\scriptsize{\makecell{Fusion\\block}}}
				& \multicolumn{1}{c}{\scriptsize{\makecell{Reg.\\head}}}
				&  \\
				\midrule
				- & \checkmark\tnote{*} & - & - & {\scriptsize{cla}} & 0.51 \\
				\checkmark\tnote{*} & - & - & - & {\scriptsize{cla}} & 0.60 \\
				\checkmark & \checkmark & - & \checkmark & {\scriptsize{cla}} & 0.57 \\
				\checkmark & \checkmark & \checkmark & - & {\scriptsize{cla}} & 0.62 \\
				\checkmark\tnote{$\S$} & \checkmark & \checkmark & \checkmark & {\scriptsize{cla}} & 0.59 \\
				\checkmark\tnote & \checkmark & \checkmark & \checkmark & {\scriptsize{reg}} & 0.49 \\
				\checkmark\tnote & \checkmark & \checkmark & \checkmark & {\scriptsize{cla}} & 0.64 \\
				\bottomrule
			\end{tabular}
			\begin{tablenotes}
				\item[*] For fairness, channels of corresponding models are adjusted to ensure that the model size is similar to others.
				\item[$\S$] The difference calculated by subtracting two fingerprint images is input into this branch instead of phase in other groups.
			\end{tablenotes}
		\end{threeparttable}
	\end{center}
\end{table}

\begin{table}[!t]
	\caption{Ablation Study of the Proposed Network with Different Numbers of Stack Stages of Feature Interaction Module on Hisign MPF}
	\label{tab:ablation_number}
	\vspace{-0.4cm}
	\begin{center}
		\begin{threeparttable}
			\begin{tabular}{*{1}{p{.25\linewidth}<{\raggedright}}*{5}{p{0.045\linewidth}<{\centering}}}
				\toprule
				{\scriptsize{\textbf{Stacking stages}}} & 2 & 3 & 4 & 5 & 6\\
				\midrule
				{\scriptsize{\textbf{NCC}}} & 0.59 & 0.62 & 0.64 & 0.63 & 0.63 \\
				{\scriptsize{\textbf{Param (M)}}} & 8.71 & 10.9 & 13.0 & 15.2 & 17.4 \\
				
				\bottomrule
			\end{tabular}
		\end{threeparttable}
	\end{center}
\end{table}

\subsection{Ablation Study}
      
We perform ablation studies on \emph{Hisign MPF} and use Equation \ref{eq:ncc} to examine the performance in registration accuracy.
Table \ref{tab:ablation_structure} presents the experiment results of specific modules and strategies in our proposed network, where ``cla'' and ``reg'' denote the construction of registration head in form of classification (introduced in Section \ref{subsubsec:registration_estimation}) or regression (same as previous works \cite{cui2019dense},\cite{cui2021dense}). The information exchange part between two branches in the feature interaction stage is called ``information interaction'', and ASPP used in the registration estimation stage for improving compatibility is represented by ``fusion block''.
These comparisons first verify the effectiveness of correlation information compared to directly extracting texture feature without constraints, while proving that integrating the two can achieve better performance.
In particular, using phase as correlation information is significantly superior to image difference because phase contains richer information about the direction and value of displacement at each pixel \cite{cui2018phase}.
The results also strongly validate the positive effects of specific designs in our network, including dual semantic feature fusion strategy, corresponding auxiliary module ASPP and the decision of output format.
In our proposed network, the stack number of information interaction module can be freely adjusted according to actual data scale. 
Table \ref{tab:ablation_number} shows corresponding ablation results under the data protocol of this paper, where the accuracy keeps improving with more stacking stages until the model overfits after $4$ steps.

\subsection{Efficiency Analysis}

\begin{table}[!t]
	\caption{Model Size and Average Time Cost of Different Fingerprint Registration Algorithms for Processing a $640 \times 640$ fingerprint pair in Hisign MPF}
	\label{tab:efficiency}
	\vspace{-0.4cm}
	\begin{center}
		\begin{threeparttable}
			\begin{tabular}{*{1}{p{.32\linewidth}<{\raggedright}}*{2}{p{0.15\linewidth}<{\centering}}}
				\toprule
				{\scriptsize{\textbf{Method}}} 
				& {\scriptsize{\textbf{Param (M)}}}
				& {\scriptsize{\textbf{Time (s)}}} \\
				\midrule
				\multirow{1}{*}{TPS Based} 
				& - & 0.41 \\
				\multirow{1}{*}{Phase Based \cite{cui2018phase}}
				& - & 21.31 \\
				\multirow{1}{*}{DRN (local)\tnote{\dag}\;\cite{cui2019dense}}
				& 0.2 & 1.17 \\
				\multirow{1}{*}{DRN (global)\tnote{\dag}\;\cite{cui2021dense}}
				& 65.0 & 0.56 \\
				\midrule
				\multirow{1}{*}{Proposed\tnote{\dag}}
				& 13.0 & 0.43 \\
				\bottomrule
			\end{tabular}
		\begin{tablenotes}
			\item[\dag] Registration method based on neural networks.
		\end{tablenotes}
		\end{threeparttable}
	\end{center}
\end{table}

Model size and inference speed of different fingerprint registration algorithms on \emph{Hisign MPF} are listed in Table \ref{tab:efficiency}.
The time covers the process from inputting a pair of $640 \times 640$ fingerprints to outputting the corresponding deformation field, which is measured on a single NVIDIA GeForce RTX 3090 GPU by setting the batch size to 1, with a 2.4 GHz CPU. 
All algorithms are implemented in Python.
It can be seen that all deep learning based methods are significantly faster than traditional methods \cite{cui2018phase}.
As can be observed from Table \ref{tab:efficiency} and Section \ref{subsec:registration_accuracy}, \ref{subsec:matching_performance}, our proposed fingerprint registration algorithm is quite competitive in efficiency  while leading in accuracy and robustness.

\section{Conclusion}\label{sec:conclusion}
In this paper, we propose a fingerprint registration algorithm to estimate the dense deformation field between two fingerprints. 
Both high-resolution phase information extracted by Gabor filters and low-resolution texture information extracted by convolutional layers are utilized in our proposed network. 
A multi-stage dual-branch information interaction mechanism is introduced to aggregate features of these two semantics at different resolutions.
Moreover, we use discrete classification in the output header instead of the previous direct regression, which implicitly introducing the correlation between numerical distributions.
Extensive experiments demonstrate that our method surpasses state-of-the-art fingerprint registration algorithms in accuracy and robustness, while also exhibiting notable efficiency advantages.
On the other hand, current solution are still not ideal in some extreme scenarios where fingerprints are severely contaminated or barely overlap.
In the future, we will further explore the design of feature extraction and multi-information fusion to overcome these difficulties.


{
	\bibliographystyle{IEEEtran}
	\bibliography{egbib}{}
}

\end{document}